\documentclass[twoside]{article}

\usepackage[accepted]{aistats2023}

\usepackage[utf8]{inputenc} 
\usepackage[T1]{fontenc}    
\usepackage{nameref,zref-xr}
\usepackage[colorlinks,linkcolor=magenta,citecolor=blue, pagebackref=true]{hyperref}     
\usepackage{url}            
\usepackage{booktabs}       
\usepackage{amsfonts}       
\usepackage{nicefrac}       
\usepackage{microtype}      
\usepackage{xcolor}         
\usepackage{array}
\usepackage{float}
\usepackage{textcomp}
\usepackage{multirow}
\usepackage{amsmath}
\usepackage{amsthm}
\usepackage{amssymb}
\usepackage{stmaryrd}
\usepackage{graphicx} 
\usepackage{bmpsize}
 \usepackage{bmpsize}
\usepackage{xargs}[2008/03/08]
\usepackage{algorithmic}
\usepackage{caption}
\usepackage{algorithm}
\usepackage{wrapfig}
\usepackage{subcaption}

\usepackage{amsmath,amsfonts,bm}

\def\eqref#1{equation~\ref{#1}}

\def\1{\bm{1}}

\def\vt{{\bm{t}}}

\def\mA{{\bm{A}}}
\def\mB{{\bm{B}}}

\def\mF{{\bm{F}}}

\def\mH{{\bm{H}}}

\def\mK{{\bm{K}}}

\def\mS{{\bm{S}}}
\def\mT{{\bm{T}}}

\def\mX{{\bm{X}}}

\def\mZ{{\bm{Z}}}

\DeclareMathAlphabet{\mathsfit}{\encodingdefault}{\sfdefault}{m}{sl}
\SetMathAlphabet{\mathsfit}{bold}{\encodingdefault}{\sfdefault}{bx}{n}

\newcommand{\KL}{D_{\mathrm{KL}}}
\newcommand{\Var}{\mathrm{Var}}

\newcommand{\Cov}{\mathrm{Cov}}

\DeclareMathOperator*{\argmax}{arg\,max}
\DeclareMathOperator*{\argmin}{arg\,min}

\newtheorem{thm}{Theorem}[section]

\newtheorem{lem}[thm]{Lemma}

\newcommand{\noun}[1]{\textsc{#1}}
\providecommand{\tabularnewline}{\\}
\floatstyle{ruled}
\newfloat{algorithm}{tbp}{loa}
\providecommand{\algorithmname}{Algorithm}
\floatname{algorithm}{\protect\algorithmname}

\usepackage[normalem]{ulem}

\usepackage[round]{natbib}   
\begin{document}

%

%

\twocolumn[

\aistatstitle{Global-Local Regularization Via Distributional Robustness}

\aistatsauthor{ Hoang Phan$^{\diamond}$ \hspace*{11mm} Trung Le$^{\dagger}$ \hspace*{11mm} Trung Phung$^{+}$ \hspace*{11mm} Anh Bui$^{\dagger}$ \hspace*{11mm} Nhat Ho$^{\ddagger}$ \hspace*{11mm} Dinh Phung$^{\dagger}$   }

\aistatsaddress{   \hspace*{-8mm} VinAI Research$^{\diamond}$  \And  \hspace*{-8mm}  Monash University, Australia$^{\dagger}$ \And   \hspace*{-8mm}  Johns Hopkins University$^{+}$    \And \hspace*{-8mm} University of Texas, Austin$^\ddagger$ } ]

\global\long\def\sidenote#1{\marginpar{\small\emph{{\color{Medium}#1}}}}%

\global\long\def\se{\hat{\text{se}}}%
\global\long\def\interior{\text{int}}%
\global\long\def\boundary{\text{bd}}%
\global\long\def\ML{\textsf{ML}}%
\global\long\def\GML{\mathsf{GML}}%
\global\long\def\HMM{\mathsf{HMM}}%
\global\long\def\support{\text{supp}}%
\global\long\def\new{\text{*}}%
\global\long\def\stir{\text{Stirl}}%
\global\long\def\mA{\mathcal{A}}%
\global\long\def\mB{\mathcal{B}}%
\global\long\def\expect{\mathbb{E}}%
\global\long\def\mF{\mathcal{F}}%
\global\long\def\mK{\mathcal{K}}%
\global\long\def\mH{\mathcal{H}}%
\global\long\def\mX{\mathcal{X}}%
\global\long\def\mZ{\mathcal{Z}}%
\global\long\def\mS{\mathcal{S}}%
\global\long\def\Ical{\mathcal{I}}%
\global\long\def\mT{\mathcal{T}}%
\global\long\def\Pcal{\mathcal{P}}%
\global\long\def\dist{d}%
\global\long\def\HX{\entro\left(X\right)}%
\global\long\def\entropyX{\HX}%
\global\long\def\HY{\entro\left(Y\right)}%
\global\long\def\entropyY{\HY}%
\global\long\def\HXY{\entro\left(X,Y\right)}%
\global\long\def\entropyXY{\HXY}%
\global\long\def\mutualXY{\mutual\left(X;Y\right)}%
\global\long\def\mutinfoXY{\mutualXY}%
\global\long\def\given{\mid}%
\global\long\def\gv{\given}%
\global\long\def\goto{\rightarrow}%
\global\long\def\asgoto{\stackrel{a.s.}{\longrightarrow}}%
\global\long\def\pgoto{\stackrel{p}{\longrightarrow}}%
\global\long\def\dgoto{\stackrel{d}{\longrightarrow}}%
\global\long\def\lik{\mathcal{L}}%
\global\long\def\logll{\mathit{l}}%
\global\long\def\bigcdot{\raisebox{-0.5ex}{\scalebox{1.5}{\ensuremath{\cdot}}}}%
\global\long\def\sig{\textrm{sig}}%
\global\long\def\likelihood{\mathcal{L}}%
\global\long\def\vectorize#1{\mathbf{#1}}%

\global\long\def\vt#1{\mathbf{#1}}%
\global\long\def\gvt#1{\boldsymbol{#1}}%
\global\long\def\idp{\ \bot\negthickspace\negthickspace\bot\ }%
\global\long\def\cdp{\idp}%
\global\long\def\das{}%
\global\long\def\id{\mathbb{I}}%
\global\long\def\idarg#1#2{\id\left\{  #1,#2\right\}  }%
\global\long\def\iid{\stackrel{\text{iid}}{\sim}}%
\global\long\def\bzero{\vt 0}%
\global\long\def\bone{\mathbf{1}}%
\global\long\def\a{\mathrm{a}}%
\global\long\def\ba{\mathbf{a}}%
\global\long\def\b{\mathrm{b}}%
\global\long\def\bb{\mathbf{b}}%
\global\long\def\B{\mathrm{B}}%
\global\long\def\boldm{\boldsymbol{m}}%
\global\long\def\c{\mathrm{c}}%
\global\long\def\C{\mathrm{C}}%
\global\long\def\d{\mathrm{d}}%
\global\long\def\D{\mathrm{D}}%
\global\long\def\N{\mathrm{N}}%
\global\long\def\h{\mathrm{h}}%
\global\long\def\H{\mathrm{H}}%
\global\long\def\bH{\mathbf{H}}%
\global\long\def\K{\mathrm{K}}%
\global\long\def\M{\mathrm{M}}%
\global\long\def\bff{\vt f}%
\global\long\def\bx{\mathbf{\mathbf{x}}}%

\global\long\def\bl{\boldsymbol{l}}%
\global\long\def\s{\mathrm{s}}%
\global\long\def\T{\mathrm{T}}%
\global\long\def\bu{\mathbf{u}}%
\global\long\def\v{\mathrm{v}}%
\global\long\def\bv{\mathbf{v}}%
\global\long\def\bo{\boldsymbol{o}}%
\global\long\def\bh{\mathbf{h}}%
\global\long\def\bs{\boldsymbol{s}}%
\global\long\def\x{\mathrm{x}}%
\global\long\def\bx{\mathbf{x}}%
\global\long\def\bz{\mathbf{z}}%
\global\long\def\hbz{\hat{\bz}}%
\global\long\def\z{\mathrm{z}}%
\global\long\def\y{\mathrm{y}}%
\global\long\def\bxnew{\boldsymbol{y}}%
\global\long\def\bX{\boldsymbol{X}}%
\global\long\def\tbx{\tilde{\bx}}%
\global\long\def\by{\mathbf{y}}%
\global\long\def\bY{\boldsymbol{Y}}%
\global\long\def\bZ{\boldsymbol{Z}}%
\global\long\def\bU{\boldsymbol{U}}%
\global\long\def\bn{\boldsymbol{n}}%
\global\long\def\bV{\boldsymbol{V}}%
\global\long\def\bI{\boldsymbol{I}}%
\global\long\def\J{\mathrm{J}}%
\global\long\def\bJ{\mathbf{J}}%
\global\long\def\w{\mathrm{w}}%
\global\long\def\bw{\vt w}%
\global\long\def\bW{\mathbf{W}}%
\global\long\def\balpha{\gvt{\alpha}}%
\global\long\def\bdelta{\boldsymbol{\delta}}%
\global\long\def\bsigma{\gvt{\sigma}}%
\global\long\def\bbeta{\gvt{\beta}}%
\global\long\def\bmu{\gvt{\mu}}%
\global\long\def\btheta{\boldsymbol{\theta}}%
\global\long\def\blambda{\boldsymbol{\lambda}}%
\global\long\def\bgamma{\boldsymbol{\gamma}}%
\global\long\def\bpsi{\boldsymbol{\psi}}%
\global\long\def\bphi{\boldsymbol{\phi}}%
\global\long\def\bpi{\boldsymbol{\pi}}%
\global\long\def\bomega{\boldsymbol{\omega}}%
\global\long\def\bepsilon{\boldsymbol{\epsilon}}%
\global\long\def\btau{\boldsymbol{\tau}}%
\global\long\def\bxi{\boldsymbol{\xi}}%
\global\long\def\realset{\mathbb{R}}%
\global\long\def\realn{\realset^{n}}%
\global\long\def\integerset{\mathbb{Z}}%
\global\long\def\natset{\integerset}%
\global\long\def\integer{\integerset}%

\global\long\def\natn{\natset^{n}}%
\global\long\def\rational{\mathbb{Q}}%
\global\long\def\rationaln{\rational^{n}}%
\global\long\def\complexset{\mathbb{C}}%
\global\long\def\comp{\complexset}%

\global\long\def\compl#1{#1^{\text{c}}}%
\global\long\def\and{\cap}%
\global\long\def\compn{\comp^{n}}%
\global\long\def\comb#1#2{\left({#1\atop #2}\right) }%
\global\long\def\param{\vt w}%
\global\long\def\Param{\Theta}%
\global\long\def\meanparam{\gvt{\mu}}%
\global\long\def\Meanparam{\mathcal{M}}%
\global\long\def\meanmap{\mathbf{m}}%
\global\long\def\logpart{A}%
\global\long\def\simplex{\Delta}%
\global\long\def\simplexn{\simplex^{n}}%
\global\long\def\dirproc{\text{DP}}%
\global\long\def\ggproc{\text{GG}}%
\global\long\def\DP{\text{DP}}%
\global\long\def\ndp{\text{nDP}}%
\global\long\def\hdp{\text{HDP}}%
\global\long\def\gempdf{\text{GEM}}%
\global\long\def\rfs{\text{RFS}}%
\global\long\def\bernrfs{\text{BernoulliRFS}}%
\global\long\def\poissrfs{\text{PoissonRFS}}%
\global\long\def\grad{\gradient}%
\global\long\def\gradient{\nabla}%
\global\long\def\partdev#1#2{\partialdev{#1}{#2}}%
\global\long\def\partialdev#1#2{\frac{\partial#1}{\partial#2}}%
\global\long\def\partddev#1#2{\partialdevdev{#1}{#2}}%
\global\long\def\partialdevdev#1#2{\frac{\partial^{2}#1}{\partial#2\partial#2^{\top}}}%
\global\long\def\closure{\text{cl}}%
\global\long\def\cpr#1#2{\Pr\left(#1\ |\ #2\right)}%
\global\long\def\var{\text{Var}}%
\global\long\def\Var#1{\text{Var}\left[#1\right]}%
\global\long\def\cov{\text{Cov}}%
\global\long\def\Cov#1{\cov\left[ #1 \right]}%
\global\long\def\COV#1#2{\underset{#2}{\cov}\left[ #1 \right]}%
\global\long\def\corr{\text{Corr}}%
\global\long\def\sst{\text{T}}%
\global\long\def\SST{\sst}%
\global\long\def\ess{\mathbb{E}}%
\newcommandx\ESS[2][usedefault, addprefix=\global, 1=]{\underset{#2}{\ess}\left[#1\right]}%
\global\long\def\Ess#1{\ess\left[#1\right]}%
\global\long\def\fisher{\mathcal{F}}%

\global\long\def\bfield{\mathcal{B}}%
\global\long\def\borel{\mathcal{B}}%
\global\long\def\bernpdf{\text{Bernoulli}}%
\global\long\def\betapdf{\text{Beta}}%
\global\long\def\dirpdf{\text{Dir}}%
\global\long\def\gammapdf{\text{Gamma}}%
\global\long\def\gaussden#1#2{\text{Normal}\left(#1, #2 \right) }%
\global\long\def\gauss{\mathbf{N}}%
\global\long\def\gausspdf#1#2#3{\text{Normal}\left( #1 \lcabra{#2, #3}\right) }%
\global\long\def\multpdf{\text{Mult}}%
\global\long\def\poiss{\text{Pois}}%
\global\long\def\poissonpdf{\text{Poisson}}%
\global\long\def\pgpdf{\text{PG}}%
\global\long\def\wshpdf{\text{Wish}}%
\global\long\def\iwshpdf{\text{InvWish}}%
\global\long\def\nwpdf{\text{NW}}%
\global\long\def\niwpdf{\text{NIW}}%
\global\long\def\studentpdf{\text{Student}}%
\global\long\def\unipdf{\text{Uni}}%
\global\long\def\transp#1{\transpose{#1}}%
\global\long\def\transpose#1{#1^{\mathsf{T}}}%
\global\long\def\mgt{\succ}%
\global\long\def\mge{\succeq}%
\global\long\def\idenmat{\mathbf{I}}%
\global\long\def\trace{\mathrm{tr}}%
\global\long\def\argmax#1{\underset{_{#1}}{\text{argmax}} }%
\global\long\def\argmin#1{\underset{_{#1}}{\text{argmin}\ } }%
\global\long\def\diag{\text{diag}}%
\global\long\def\norm{}%
\global\long\def\spn{\text{span}}%
\global\long\def\vtspace{\mathcal{V}}%
\global\long\def\field{\mathcal{F}}%
\global\long\def\ffield{\mathcal{F}}%
\global\long\def\inner#1#2{\left\langle #1,#2\right\rangle }%
\global\long\def\iprod#1#2{\inner{#1}{#2}}%
\global\long\def\dprod#1#2{#1 \cdot#2}%
\global\long\def\norm#1{\left\Vert #1\right\Vert }%
\global\long\def\entro{\mathbb{H}}%
\global\long\def\entropy{\mathbb{H}}%
\global\long\def\Entro#1{\entro\left[#1\right]}%
\global\long\def\Entropy#1{\Entro{#1}}%
\global\long\def\mutinfo{\mathbb{I}}%
\global\long\def\relH{\mathit{D}}%
\global\long\def\reldiv#1#2{\relH\left(#1||#2\right)}%
\global\long\def\KL{KL}%
\global\long\def\KLdiv#1#2{\KL\left(#1\parallel#2\right)}%
\global\long\def\KLdivergence#1#2{\KL\left(#1\ \parallel\ #2\right)}%
\global\long\def\crossH{\mathcal{C}}%
\global\long\def\crossentropy{\mathcal{C}}%
\global\long\def\crossHxy#1#2{\crossentropy\left(#1\parallel#2\right)}%
\global\long\def\breg{\text{BD}}%
\global\long\def\lcabra#1{\left|#1\right.}%
\global\long\def\lbra#1{\lcabra{#1}}%
\global\long\def\rcabra#1{\left.#1\right|}%
\global\long\def\rbra#1{\rcabra{#1}}%

\begin{abstract}
Despite superior performance in many situations, deep neural networks are often vulnerable to adversarial examples and distribution shifts, limiting model generalization ability in real-world applications. To alleviate these problems, recent approaches leverage distributional robustness optimization (DRO) to find the most challenging distribution, and then minimize loss function over this most challenging distribution. Regardless of having achieved some improvements, these DRO approaches have some obvious limitations. First, they purely focus on local regularization to strengthen model robustness, missing a global regularization effect that is useful in many real-world applications (e.g., domain adaptation, domain generalization, and adversarial machine learning). Second, the loss functions in the existing DRO approaches operate in only the most challenging distribution, hence decouple with the original distribution, leading to a restrictive modeling capability. In this paper, we propose a novel regularization technique, following the veins of Wasserstein-based DRO framework. Specifically, we define a particular joint distribution and Wasserstein-based uncertainty, allowing us to couple the original and most challenging distributions for enhancing modeling capability and enabling both local and global regularizations. Empirical studies on different learning problems demonstrate that our proposed approach significantly outperforms the existing regularization approaches in various domains.

\end{abstract}

\section{Introduction}

As the Wasserstein (WS) distance is a powerful and convenient tool of measuring closeness between distributions, Wasserstein Distributional Robustness (WDR) has been one of the most widely-used variants of DR. Here we consider a generic Polish space $S$ endowed with a distribution $\mathbb{P}$. Let $r:S\to\mathbb{R}$ be a real-valued (risk) function and $c:S\times S\to \mathbb{R}_{+}$ be a cost function. Distributional robustness setting aims to find the distribution $\tilde{\mathbb{P}}$ in the vicinity of $\mathbb{P}$ and maximizes the risk in the expectation form \citep{blanchet2019quantifying, sinha2017certifying}:
\begin{equation}
\sup_{\mathbb{\tilde{P}}:\mathcal{W}_{c}\left(\mathbb{P},\mathbb{\tilde{P}}\right)<\epsilon}\underset{\tilde{Z}\sim\mathbb{\mathbb{\tilde{P}}}}{\mathbb{E}}\left[r\left(\tilde{Z}\right)\right],\label{eq:primal_form-1}
\end{equation}
where $\epsilon>0$ and $\mathcal{W}_{c}\left(\mathbb{P},\mathbb{\mathbb{\tilde{P}}}\right):=\inf_{\gamma\in\Gamma\left(\mathbb{P},\mathbb{\tilde{P}}\right)}\int cd\gamma\label{eq:asf-1}
$ denotes an optimal transport
(OT) or a WS distance
with the set of couplings $\Gamma\left(\mathbb{P},\mathbb{\mathbb{\tilde{P}}}\right)$ whose marginals are $\mathbb{P}$ and $\mathbb{\tilde{P}}$. 


Direct optimization over the set of distributions $\tilde{\mathbb{P}}$ is often computationally intractable except in limited cases, we thus seek to cast this problem into its dual form. With the assumption that $r\in L^{1}\left(\mathbb{P}\right)$ is upper semi-continuous and the cost $c$ is a non-negative and continuous function satisfying $c(Z,\tilde{Z})=0\text{ iff }Z=\tilde{Z}$, \citep{blanchet2019quantifying, sinha2017certifying} showed the \emph{dual} form for Eq. (\ref{eq:primal_form-1}) is:
\begin{equation}
   \inf_{\lambda\geq0}\left\{ \lambda\epsilon+\ESS[\sup_{\tilde{Z}}\left\{ r\left(\tilde{Z}\right)-\lambda c\left(\tilde{Z},Z\right)\right\} ]{Z\sim\mathbb{P}}\right\} .\label{eq:dual_form-1}  
\end{equation}

When applying DR to the supervised learning setting, $\tilde{Z}=\left(\tilde{X},\tilde{Y}\right)$
is a pair of data/label drawn from $\mathbb{\mathbb{\tilde{P}}}$
and $r$ is the loss function \citep{blanchet2019quantifying, sinha2017certifying}.
The fact that $r$ engages only $\tilde{Z}=\left(\tilde{X},\tilde{Y}\right)\sim\mathbb{\tilde{P}}$
certainly restricts the modeling capacity of (\ref{eq:dual_form-1}). The reasons are as follows. Firstly, for each anchor $Z$, the most challenging sample $\tilde{Z}$ is currently defined as the one maximizing $\sup_{\tilde{Z}} \left\{r(\tilde{Z}) - \lambda c(Z, \tilde{Z}) \right\}$, where $r(\tilde{Z})$ is inherited from the primal form (\ref{eq:primal_form-1}). Hence, it is not suitable to express the risk function $r$ engaging both $Z$ and $\tilde{Z}$ (e.g., Kullback-Leibler divergence $KL\left(p\left(\tilde{Z}\right)\Vert p(Z)\right)$ between the predictions for $Z$ and $\tilde{Z}$ as in TRADES \citep{zhang2019theoretically}). Secondly, it is also
\emph{impossible} to inject a\emph{ global regularization term} involving a batch of samples $\tilde{Z}$ and $Z$. 

\textbf{Contribution.} To empower the formulation of DR for efficiently  tackling
various real-world problems, in this work, we propose a rich OT
based DR framework, named \textbf{\emph{G}}\emph{lobal-}\textbf{\emph{L}}\emph{ocal
}\textbf{\emph{O}}\emph{ptimal }\textbf{\emph{T}}\emph{ransport based
}\textbf{\emph{D}}\emph{istributional }\textbf{\emph{R}}\emph{obustness}
(GLOT-DR). Specifically, by designing special joint distributions
$\mathbb{P}$ and $\tilde{\mathbb{P}}$ together with some constraints,
our framework is applicable to a mixed variety of real-world applications,
including {domain generalization} (DG),  {domain adaptation}
(DA),  {semi-supervised learning} (SSL), and {adversarial
machine learning} (AML). 

Additionally, our GLOT-DR makes it possible for us to equip not only  a \emph{local regularization term} for enforcing a local
smoothness and robustness, but also a \emph{global regularization term}
to impose a global effect targeting a downstream task. Moreover,
by designing a specific WS distance, we successfully develop a closed-form
solution for GLOT-DR without using the dual form in \citep{blanchet2019quantifying, sinha2017certifying}
(i.e., Eq. (\ref{eq:dual_form-1})). 

Technically, our solution turns
solving the inner maximization in the dual form (\ref{eq:dual_form-1})
into sampling a set of challenging particles according to a local distribution, 
on which we can handle efficiently using Stein Variational Gradient
Decent (SVGD) \citep{NIPS2016_b3ba8f1b} approximate inference algorithm. Based on the general framework
of GLOT-DR, we establish the settings for DG, DA, SSL, and AML and conduct experiments to compare our GLOT-DR to state-of-the-art baselines in these real-world applications. Overall, our contributions can be
summarized as follows:
\vspace*{-2mm}
\begin{itemize}
    \item We enrich the general framework of DR to make it possible for many real-world applications by enforcing both local and global regularization terms. We note that the global regularization term is crucial for many downstream tasks (see Section \ref{subsec:Our-Framework} for more details).
\vspace*{-1mm}
    \item We propose a closed-form solution for our GLOT-DR without involving the dual form in \citep{blanchet2019quantifying, sinha2017certifying}
(i.e., Eq. (\ref{eq:dual_form-1})). We note that the dual form (\ref{eq:dual_form-1}) is \emph{not computationally tractable} due to the minimization over $\lambda$.
\vspace*{-1mm}
    \item We conduct comprehensive experiments to compare our GLOT-DR to state-of-the-art baselines in DG, DA, SSL, and AML. The experimental results demonstrate the merits of our proposed approach and empirically prove that both of the introduced local and global regularization terms advance existing methods across various scenarios, including DG, DA, SSL, and AML.
\end{itemize}

\section{Related Work}
\textbf{Distributional robustness (DR).} DR is an attractive framework
for improving machine learning
models in terms of robustness and generalization. Its underlying idea is to find the \emph{most challenging distribution} around a given distribution and then challenge a model with this distribution. To characterize
the closeness of a distribution to a center distribution, either a $f$-divergence \citep{ben_tal,duchi2021statistics,duchi2019distributionally,miyato2015distributional,namkoong2016stochastic}
or Wasserstein distance \citep{blanchet2019robust,gao2016distributionally,kuhn2019wasserstein, mohajerin2015data,shafieezadeh2015distributionally}
can be employed. Other works \citep{blanchet2019quantifying,sinha2017certifying}
developed a dual form for DR, opening the door to incorporate DR into
the training of deep learning models.

\textbf{Adversarial Robustness (AR)}. Neural networks are generally
vulnerable to adversarial attacks, notably FGSM \citep{goodfellow2014explaining},
PGD \citep{madry2017towards}, and Auto-Attack \citep{croce2020reliable}.
Among various kinds of defense approaches, Adversarial Training (AT),
originating in \citep{goodfellow2014explaining}, has drawn the most
research attention. Given its effectiveness and efficiency, many variants
of AT have been proposed with: (1) different types of adversarial examples
(e.g., the worst-case examples \citep{goodfellow2014explaining} or
most divergent examples \citep{zhang2019theoretically}), (2) different
searching strategies (e.g., non-iterative FGSM and Rand FGSM\noun{
}\citep{madry2017towards}), (3) additional regularization (e.g., adding
constraints in the latent space \citep{bui2020improving,zhang2019defense}).
Inspired by the potential of DR, it has been applied to enhance model robustness in \citep{deng2020adversarial,levine2020wasserstein,VAT,sinha2017certifying, nguyen2022particle, bui2022unified, le22c, hoang20c}.

\textbf{Transfer Learning (TL)}. Domain adaptation (DA) and domain
generalization (DG) are two typical settings in TL. As for domain adaptation,  \citep{ganin2016domain,li2020enhanced,long2017conditional, nguyen2022cycle, le21a_lamda, nguyen21a_most, Nguyen_2021_ICCV,nguyen2021tidot}
aim at training a model based on a labeled source domain to adapt
to an unlabeled target domain, while the works in DG \citep{balaji2018metareg,bousmalis2016domain,li2017deeper,li2018learning,li2019episodic,mancini2018best, phung_dg}
aim at training a model based on multiple labeled source domains to
predict well on unseen target domains. Finally, in more recent work, it was leveraged with DG in
\citep{zhao_maximum} and DA in \citep{wang2021distributionally}. 

\begin{figure*}[t]

\centering
\includegraphics[width=.9\textwidth, trim=.0cm 2.4cm 0.5cm 1.2cm,clip]{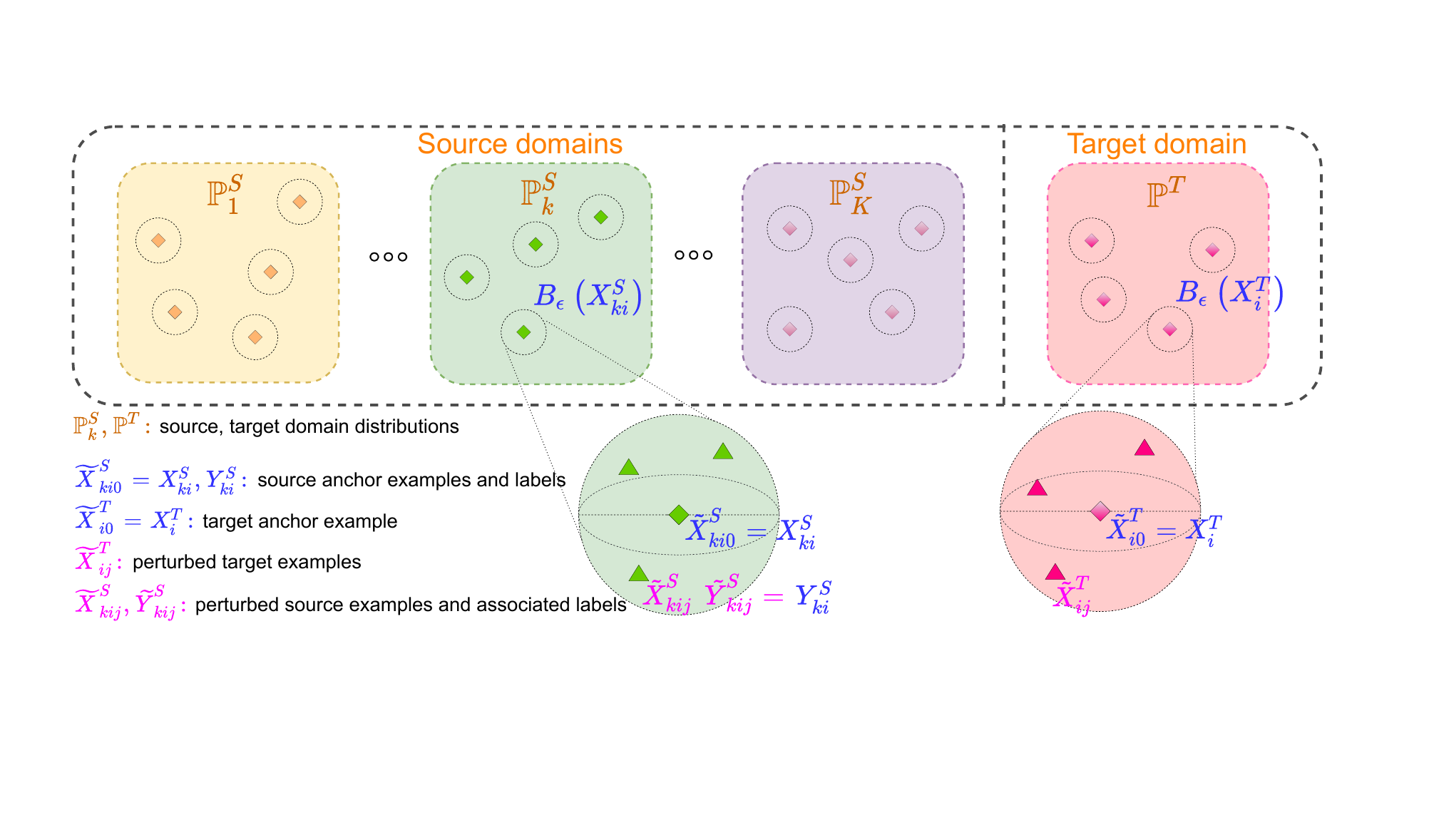}

\caption{Overview of GLOT-DR. We sample $\left[X_{ki}^{S},Y_{ki}^{S}\right]_{i=1}^{B_{k}^{S}}$
for each source domain, $\left[X_{i}^{T}\right]_{i=1}^{B^{T}}$ for
the target domain, and define $Z,\tilde{Z}$ as in Eqs. (\ref{eq:Z_construct},\ref{eq:Z_til_construct}). For $\left(Z,\tilde{Z}\right)\sim\gamma$
satisfying $\mathbb{E}_{\gamma}\left[\rho\left(Z,\tilde{Z}\right)\right]^{1/q}\protect\leq\epsilon$,
we have $\tilde{X}_{ki0}^{S}=X_{ki0}^{S}=X_{ki}^{S}$ , $\tilde{X}_{i0}^{T}=X_{i0}^{T}=X_{i}^{T}$.
Besides, $\tilde{X}_{kij}^{S}$ with $j\protect\geq1$ can be viewed
as the \emph{perturbed} examples in the ball $B_{\epsilon}\left(X_{ki}^{S}\right)$,
which have the same label $Y_{ki}^{S}$. Similarly, $\tilde{X}_{ij}^{T}$
with $j\protect\geq1$ can be viewed as the \emph{perturbed}
examples in the ball $B_{\epsilon}\left(X_{i}^{T}\right)$.\label{fig:Overview-of-GLOT-DR.}}

\end{figure*}

\section{Proposed Approach}
In this section, we first introduce the GLOT-DR framework and provide the theoretical development in Section  \ref{subsec:Our-Framework}. Then Section \ref{subsec:training-procedure} presents the general training procedure of our proposed approach, and the detailed formulations of scenarios are described in the remainder of this section.
\subsection{Our Framework \label{subsec:Our-Framework}}

We propose a regularization technique based on optimal transport distributional robustness that can be widely applied to many settings including i) \emph{semi-supervised
learning}, ii) \emph{domain adaptation}, iii) \emph{domain
generalization}, and iv) \emph{adversarial machine learning}
. In what follows, we present the general setting along with the notations used throughout the paper and technical
details of our framework. 

Assume that we have \emph{multiple labeled source domains} with the
\emph{data/label} distributions $\left\{ \mathbb{P}_{k}^{S}\right\} _{k=1}^{K}$
and a \emph{single unlabeled target domain} with the \emph{data} distribution
$\mathbb{P}^{T}$. For the $k$-th source domain, we draw a batch
of $B_{k}^{S}$ examples as $\left(X_{ki}^{S},Y_{ki}^{S}\right)\iid\mathbb{P}_{k}^{S}$,
where $i=1,\ldots,B_{k}^{S}$. Meanwhile, for
the target domain, we sample a batch of $B^{T}$ examples as $X_{i}^{T}\iid\mathbb{P}^{T},\,i=1,\ldots,B^{T}$.
It is worth noting that for the DG setting, we set $B^{T}=0$ (i.e.,
not use any target data in training). Furthermore, we examine the
multi-class classification problem with the label set $\mathcal{Y}:=\left\{ 1,...,M\right\} $.
Hence, the prediction of a classifier is a prediction probability
belonging to the \emph{label simplex} $\Delta_{M}:=\left\{ \pi\in\mathbb{R}^{M}:\norm{\pi}_{1}=1\,\text{and}\,\pi\geq\bzero\right\} $.
Finally, let $f_{\psi}=h_{\theta}\circ g_{\phi}$ with $\psi=(\phi,\theta)$
be parameters of our deep net, wherein $g_{\phi}$ is the feature
extractor and $h_{\theta}$ is the classifier on top of feature representations.


\textbf{Constructing Challenging Samples:} As explained below, our method involves the construction of a random
variable $Z$ with distribution $\mathbb{P}$ and another random variable
$\tilde{Z}$ with distribution $\mathbb{\tilde{P}}$, ``containing''
anchor samples $\left(X_{ki}^{S},Y_{ki}^{S}\right),X_{i}^{T}$ and
their perturbed counterparts $\left(\tilde{X}_{kij}^{S},\tilde{Y}_{kij}^{S}\right),\tilde{X}_{ij}^{T}$ (see Figure \ref{fig:Overview-of-GLOT-DR.} for the illustration).
The inclusion of both anchor samples and perturbed samples allows us to define a unifying cost function containing local regularization, global regularization, and classification loss.

Concretely, we first start with the construction of $Z$, containing
repeated anchor samples as follows:
\begin{align}
Z & :=\left[\left[\left[X_{kij}^{S},Y_{kij}^{S}\right]_{k=1}^{K}\right]_{i=1}^{B_{k}^{S}}\right]_{j=0}^{n^{S}},\left[\left[X_{ij}^{T}\right]_{i=1}^{B^{T}}\right]_{j=0}^{n^{T}}.\label{eq:Z_construct}
\end{align}
Here, each source sample is repeated $n^{S}+1$ times $(X_{kij}^{S},Y_{kij}^{S})=(X_{ki}^{S},Y_{ki}^{S}),\,\forall j$,
while each target sample is repeated $n^{T}+1$ times $X_{ij}^{T}=X_{i}^{T},\,\forall j$.
The corresponding distribution of this random variable is denoted
as $\mathbb{P}$.  In contrast to $Z$, we next define random variable
$\tilde{Z}\sim\tilde{\mathbb{P}}$, whose form is
\begin{equation}
\tilde{Z}:=\left[\left[\left[\tilde{X}_{kij}^{S},\tilde{Y}_{kij}^{S}\right]_{k=1}^{K}\right]_{i=1}^{B_{k}^{S}}\right]_{j=0}^{n^{S}},\left[\left[\tilde{X}_{ij}^{T}\right]_{i=1}^{B^{T}}\right]_{j=0}^{n^{T}}.\label{eq:Z_til_construct}
\end{equation}
Here we note that for $\tilde{X}^S_{kij}$, the index $k$ specifies the $k$-th source domain, the index $i$ specifies an example in the $k$-th source batch, while the index $j$ specifies the $j$-th perturbed example to the source example $X^S_{ki}$. Similarly, for $\tilde{X}^T_{ij}$, the index $i$ specifies an example in the target batch, while the index $j$ specifies the $j$-the perturbed example to the target example $X^T_i$. 

We would like $\tilde{Z}$ to contain both: i) anchor examples, i.e.,
$\left(\tilde{X}_{ki0}^{S},\tilde{Y}_{ki0}^{S}\right)=\left(X_{ki}^{S},Y_{ki}^{S}\right)$
and $\tilde{X}_{i0}^{T}=X_{i}^{T}$; ii) $n^{S}$ perturbed source
samples $\left\{ \left(\tilde{X}_{kij}^{S},\tilde{Y}_{kij}^{S}\right)\right\} _{j=1}^{n^{S}}$ to $\left(X^S_{ki},Y^S_{ki}\right)$
and $n^{T}$ perturbed target samples $\left\{ \tilde{X}_{ij}^{T}\right\} _{i=1}^{n^{T}}$ to $X^T_i$.
In order to impose this requirement, we only consider sampling $\tilde{Z}$
from distribution $\tilde{\mathbb{P}}$ inside the Wasserstein-ball
of $\mathbb{P}$, i.e., satisfying $\mathcal{W}_{\rho}\left(\mathbb{P},\tilde{\mathbb{P}}\right):=\underset{\gamma\in\Gamma\left(\mathbb{P},\tilde{\mathbb{P}}\right)}{\inf}\underset{\left(Z,\tilde{Z}\right)\sim\gamma}{\mathbb{E}}\left[\rho\left(Z,\tilde{Z}\right)\right]^{\frac{1}{q}}\leq\epsilon$,
where the cost metric $\rho$ is defined as
\begin{multline*}
\rho\left(Z,\tilde{Z}\right) 
 :=\infty\sum_{k=1}^{K}\sum_{i=1}^{B_{k}^{S}}\norm{X_{ki0}^{S}-\tilde{X}_{ki0}^{S}}_{p}^{q}
\\  + \infty\sum_{i=1}^{B^{T}}\norm{X_{i0}^{T}-\tilde{X}_{i0}^{T}}_{p}^{q}  + \sum_{k=1}^{K}\sum_{i=1}^{B_{k}^{S}}\sum_{j=1}^{n^{S}}\norm{X_{kij}^{S}-\tilde{X}_{kij}^{S}}_{p}^{q} 
\\ + \sum_{i=1}^{B^{T}}\sum_{j=1}^{n^{T}}\norm{X_{ij}^{T}-\tilde{X}_{ij}^{T}}_{p}^{q}  +\infty\sum_{k=1}^{K}\sum_{i=1}^{B_{k}^{S}}\sum_{j=0}^{n^{S}}\rho_{l}\left(Y_{kij}^{S},\tilde{Y}_{kij}^{S}\right),
\end{multline*}
where $\rho_{l}$ is a metric on the \emph{label simplex} $\Delta_{M}$ and $q \geq 1$.
Here we slightly abuse the notion by using $Y\in\mathcal{Y}$ to represent
its corresponding one-hot vector.  By definition, this cost metric
almost surely: i) enforces all $0$-th (i.e., $j=0$) samples in $\tilde{Z}$ to be anchor samples, i.e., $\tilde{X}_{ki0}^{S}=X_{ki0}= X_{ki}$; ii) allows perturbations on the input data, i.e.,
$\tilde{X}_{kij}^{S}\neq X_{ki}^{S}$ and $\tilde{X}_{ij}^{T}\neq X_{i}^{T}$,
for $\forall j\neq0$; iii) restricts perturbations on labels, i.e.,
$Y_{kij}^{S}=\tilde{Y}_{kij}^{S}$ for $\forall j$ (see Figure  \ref{fig:Overview-of-GLOT-DR.} for the illustration). The reason is that if either (i) or (iii) is violated on a non-zero measurable set then $\mathcal{W}_{\rho}\left(\mathbb{P},\tilde{\mathbb{P}}\right)$ becomes infinity.


\textbf{Learning Robust Classifier:}
Upon clear definitions of $\tilde{Z}$ and $\tilde{\mathbb{P}}$,
we wish to learn good representations and regularize the classifier
$f_{\psi}$, via the following DR problem:
\begin{equation}
\min_{\theta,\phi}\max_{\tilde{\mathbb{P}}:\mathcal{W}_{\rho}\left(\mathbb{P},\tilde{\mathbb{P}}\right)\leq\epsilon}\mathbb{E}_{\tilde{Z}\sim\tilde{\mathbb{P}}}\left[r\left(\tilde{Z};\phi,\theta\right)\right].\label{eq:dro_p1}
\end{equation}
The cost function $r\left(\tilde{Z};\phi,\theta\right):=\alpha r^{l}\left(\tilde{Z};\phi,\theta\right)+\beta r^{g}\left(\tilde{Z};\phi,\theta\right)+\mathcal{L}\left(\tilde{Z};\phi,\theta\right)$
with $\alpha,\beta>0$ is defined as the weighted sum of a \emph{local-regularization
function} $r^{l}\left(\tilde{Z};\phi,\theta\right)$, a \emph{global-regularization
function} $r^{g}\left(\tilde{Z};\phi,\theta\right)$, and the \emph{loss
function} $\mathcal{L}\left(\tilde{Z};\phi,\theta\right)$, whose
explicit forms are dependent on the task (DA, SSL, DG, and AML).

Intuitively, the optimization in Eq.  (\ref{eq:dro_p1}) iteratively searches
for the worst-case $\tilde{\mathbb{P}}$ w.r.t. the cost $r\left(\cdot;\phi,\theta\right)$,
then changes the network $f_{\psi}$ to minimize the worst-case cost.

We now define 
$$
\Gamma_{\epsilon}:=\left\{ \gamma:\gamma\in\underset{\tilde{\mathbb{P}}}{\bigcup} \;\Gamma\left(\mathbb{P},\tilde{\mathbb{P}}\right) ,\underset{\left(Z,\tilde{Z}\right)\sim\gamma}{\mathbb{E}}\left[\rho\left(Z,\tilde{Z}\right)\right]^{1/q}\leq\epsilon\right\}
$$
 and show that the inner max problem in Eq. (\ref{eq:dro_p1}) is equivalent
to searching in $\Gamma_{\epsilon}$.

\begin{lem}
\label{lem:dro_p2} The optimization
problem in Eq. (\ref{eq:dro_p1}) is equivalent to the following optimization
problem:
\begin{equation}
\min_{\theta,\phi}\max_{\gamma:\in\Gamma_{\epsilon}}\underset{\left(Z,\tilde{Z}\right)\sim\gamma}{\mathbb{E}}\left[r\left(\tilde{Z};\phi,\theta\right)\right].\label{eq:dro_p2}
\end{equation}
\end{lem}

To tackle the optimization problem (OP) in Eq. (\ref{eq:dro_p2}), we
add the entropic regularization and arrive at the following OP:
\begin{equation}
\min_{\theta,\phi}\max_{\gamma:\in\Gamma_{\epsilon}}\left\{ \underset{\left(Z,\tilde{Z}\right)\sim\gamma}{\mathbb{E}}\left[r\left(\tilde{Z};\phi,\theta\right)\right]+\frac{1}{\lambda}\mathbb{H}\left(\gamma\right)\right\} ,\label{eq:dro_entropic}
\end{equation}
where $\lambda>0$ is the entropic regularization parameter and $\mathbb{H}$
returns the entropy of a given distribution. 

It is worth noting that minimizing the entropy $\mathbb{H}\left(\gamma\right)$ encourages more uniform $\gamma$. Moreover, when $\lambda$ becomes bigger, the optimal solution of the OP in Eq. (\ref{eq:dro_entropic}) gets closer to that of (\ref{eq:dro_p2}). Additionally, the following theorem indicates the optimal solution of the inner
max in the OP in Eq. (\ref{eq:dro_entropic}).
\begin{thm}
\label{thm:opt_sol} Assuming
$r\left(\tilde{Z};\psi\right)=\alpha r^{l}\left(\tilde{Z};\psi\right)+\beta r^{g}\left(\tilde{Z};\psi\right)+\mathcal{L}\left(\tilde{Z};\psi\right)$
with $\psi=(\phi,\theta)$. In addition, $Z$ and $\tilde{Z}$ are
constructed as in Eq.(\ref{eq:Z_construct}) and Eq.(\ref{eq:Z_til_construct}),
respectively. Let $\ell$ denote the loss function, so the expected
classification loss becomes
\[
\mathcal{L}\left(\tilde{Z};\psi\right):=\sum_{k=1}^{K}\sum_{i=1}^{B_{k}^{S}}\sum_{j=0}^{n_{k}^{S}}\ell\left(\tilde{X}_{kij}^{S},\tilde{Y}_{kij}^{S};\psi\right).
\]
Moreover, let the global-regulazation $r^{g}\left(\tilde{Z};\psi\right):=r^{g}\left(\left[\tilde{X}_{ki0}^{S}\right]_{k,i},\left[\tilde{X}_{i0}^{T}\right]_{i};\psi\right)$
depend only on anchor samples, while the local-regularization depend
on the differences between anchor samples and perturbed samples,
\begin{align*}
r^{l}\left(\tilde{Z};\psi\right)  &:=\sum_{i=1}^{B^{T}}\sum_{j=1}^{n^{T}}s\left(\tilde{X}_{i0}^{T},\tilde{X}_{ij}^{T};\psi\right)
+ \\&\sum_{k=1}^{K}\sum_{i=1}^{B_{k}^{S}}\sum_{j=1}^{n_{k}^{S}}s\left(\tilde{X}_{ki0}^{S},\tilde{X}_{kij}^{S};\psi\right),
\end{align*}
where $s\left(\tilde{X}_{0},\tilde{X}_{j};\psi\right)$ measures the
difference between 2 input samples, and $s\left(X,X;\psi\right)=0,\forall X$.
To this end, the inner max in the OP when $q=\infty$ has the following
solution{\small{}
\begin{alignat}{1}
\gamma^{*}\left(Z,\tilde{Z}\right) =\prod_{k=1}^{K}\prod_{i=1}^{B_{k}^{S}}\prod_{j=0}^{n_{k}^{S}}p_{k}^{S}\left(X_{ki}^{S},Y_{ki}^{S}\right)\prod_{i=1}^{B^{T}}\prod_{j=0}^{n^{T}}p^{T}\left(X_{i}^{T}\right)\nonumber \\
 \prod_{k=1}^{K}\prod_{i=1}^{B_{k}^{S}}\prod_{j=0}^{n_{k}^{S}}q_{ki}^{S}\left(\tilde{X}_{kij}^{S}\mid X_{ki}^{S},Y_{ki}^{S};\psi\right)  \prod_{i=1}^{B^{T}}\prod_{j=1}^{n^{T}}q_{i}^{T}\left(\tilde{X}_{ij}^{T}\mid X_{i}^{T};\psi\right),\label{eq:opt_gamma}
\end{alignat}
}where $B_{\epsilon}\left(X\right):=\left\{ X':\norm{X'-X}_{p}\leq\epsilon\right\} $
is the $\epsilon$-ball around $X$, $\left(X_{ki}^{S},Y_{ki}^{S}\right)_{i=1}^{B_{k}^{S}}\iid\mathbb{P}_{k}^{S},\forall k$,
$X_{1:B^{T}}^{T}\iid\mathbb{P}^{T}$, $p_{k}^{S}$ is the density
function of $\mathbb{P}_{k}^{S}$, $p^{T}$ is the density function
of $\mathbb{P^{T}}$, ${ q_{ki}^{S}\left(\tilde{X}_{kij}^{S}\mid X_{ki}^{S},Y_{ki}^{S};\psi\right)} \propto {\exp\left\{ \lambda[\alpha s(X_{ki}^{S},\tilde{X}_{kij}^{S};\psi)+\ell(\tilde{X}_{kij}^{S},Y_{ki}^{S};\psi)]\right\}}$
is \textbf{the local distribution over $B_{\epsilon}\left(X_{ki}^{S}\right)$} around the anchor example $X_{ki}^{S}$, and ${q_{i}^{T}\left(\tilde{X}_{ij}^{T}\mid X_{i}^{T};\psi\right)}\propto {\exp\left\{ \lambda\alpha s\left(X_{i}^{T},\tilde{X}_{ij}^{T};\psi\right)\right\}} $
is \textbf{the local distribution over} $B_{\epsilon}\left(X_{i}^{T}\right)$
around the anchor example $X_{i}^{T}$.
\end{thm}

The optimal $\gamma^*$ in Eq. (\ref{eq:opt_gamma}) involves the local distributions $q^S_{ki}$ around the anchor example $X^S_{ki}$ and $q^T_{i}$ around the anchor example $X^T_i$. By substituting the optimal solution in Eq. (\ref{eq:opt_gamma})
back to Eq. (\ref{eq:dro_p2}), we reach the following OP with $\psi=(\phi,\theta)$:
\begin{align}
\min_{\psi}\underset{\forall k:\left(X_{ki}^{S},Y_{ki}^{S}\right)_{i=1}^{B_{k}^{S}}\iid\mathbb{P}_{k}^{S},X_{1:B^{T}}^{T}\iid\mathbb{P}^{T}}{\mathbb{E}}\left[r\left(\tilde{Z};\psi\right)\right],\label{eq:last_opt}
\end{align}
where $r\left(\tilde{Z};\psi\right)$ is defined as 
\begin{align}
\underset{{\left[\tilde{X}_{kij}^{S}\right]_{j}  \sim \text{\ensuremath{q_{ki}^{S}}}}}{\mathbb{E}}\left[\alpha s(X_{ki}^{S},\tilde{X}_{kij}^{S};\psi)+\ell(\tilde{X}_{kij}^{S},Y_{ki}^{S};\psi)\right]\nonumber
\\ +  \underset{\left[\tilde{X}_{ij}^{T}\right]_{j}\sim\text{\ensuremath{q_{i}^{T}}}}{\mathbb{E}}\left[\alpha  s\left(X_{i}^{T},\tilde{X}_{ij}^{T};\psi\right)\right] \hspace*{11mm}\nonumber
\\+\beta r^{g}\left(\left[X_{ki}^{S}\right]_{k,i},\left[X_{i}^{T}\right]_{i};\psi\right)\label{eq:final_opt}
\end{align}
with the \emph{local distribution} $\ensuremath{q_{ki}^{S}}$
over $B_{\epsilon}\left(X_{ki}^{S}\right)$ 
and the \emph{local distribution} $\ensuremath{q_{i}^{T}}$
over $B_{\epsilon}\left(X_{i}^{T}\right)$. 

As shown in Eq. (\ref{eq:final_opt}), the perturbed examples $\tilde{X}_{kij}^{S}$ are sampled from the local distribution $q^S_{ki}$ over the ball $B_{\epsilon}\left(X_{ki}^{S}\right)$, while the perturbed examples $\tilde{X}_{ij}^{T}$ are sampled from the local distribution $q_{i}^{T}$ over the ball $B_{\epsilon}\left(X_{i}^{T}\right)$. Due to the formula of $q^S_{ki}$, the perturbed examples $\tilde{X}_{kij}^{S}$ tend to reach the high-likelihood region of $q^S_{ki}$ or high-valued region for $\exp\left\{ \lambda[\alpha s(X_{ki}^{S},\tilde{X}_{kij}^{S};\psi)+\ell(\tilde{X}_{kij}^{S},Y_{ki}^{S};\psi)]\right\}$. We hence can interpret $\tilde{X}_{kij}^{S}$ as adversarial examples that maximize $\lambda[\alpha s(X_{ki}^{S},\tilde{X}_{kij}^{S};\psi)+\ell(\tilde{X}_{kij}^{S},Y_{ki}^{S};\psi)]$. Subsequently, in (\ref{eq:final_opt}), we update $\psi$ to minimize $\lambda[\alpha s(X_{ki}^{S},\tilde{X}_{kij}^{S};\psi)+\ell(\tilde{X}_{kij}^{S},Y_{ki}^{S};\psi)]$ w.r.t. the perturbed adversarial examples. Similarly, we can interpret the perturbed examples $\tilde{X}_{ij}^{T}$.      

Additionally, we can equip the global-regularization function
$r^{g}\left(\left[X_{ki}^{S}\right]_{k,i},\left[X_{i}^{T}\right]_{i};\psi\right)$
\label{potential} to suit various characteristics
for the task, e.g., bridging the distribution shift between source and target domains
in DA, between labeled and unlabeled portions in SSL, and between benign and
adversarial data examples in AML, as well as learning domain invariant features
in DG. Moreover, our global and local regularization terms
can be naturally applied to the latent space induced by the feature
extractor $g_{\phi}.$ Furthermore, the theory development for this
case is similar to that for the data space except replacing $X$ in
the data space by $g_{\phi}\left(X\right)$ in the latent space.

\subsection{Training Procedure of Our Approach \label{subsec:training-procedure}}

In what follows, we present how to solve the OP in Eq. (\ref{eq:last_opt})
efficiently. Accordingly, we first need to sample $\left(X_{ki}^{S},Y_{ki}^{S}\right)_{i=1}^{B_{k}^{S}}\iid\mathbb{P}_{k}^{S},\forall k\,\text{and}\,X_{1:B^{T}}^{T}\iid\mathbb{P}^{T}$.
For each source anchor $\left(X_{ki}^{S},Y_{ki}^{S}\right)$, we sample
$\left[\tilde{X}_{kij}^{S}\right]_{j=1}^{n^{S}}\iid\ensuremath{q_{ki}^{S}}$
in the ball $B_{\epsilon}\left(X_{ki}^{S}\right)$ with the \emph{density
function proportional} to $\exp\left\{ \lambda[\alpha s(X_{ki}^{S},\bullet;\psi)+\ell(\bullet,Y_{ki}^{S};\psi)]\right\} $.
Furthermore, for each target anchor $X_{i}^{T}$, we sample $\left[\tilde{X}_{ij}^{T}\right]_{j=1}^{n^{T}}\iid\ensuremath{q_{i}^{T}}$
in the ball $B_{\epsilon}\left(X_{i}^{T}\right)$ with the d\emph{ensity
function proportional} to $\exp$$\left\{ \lambda\alpha s\left(X_{i}^{T},\bullet;\psi\right)\right\} $.

To sample the particles from their local distributions, we use Stein
Variational Gradient Decent (SVGD) \citep{NIPS2016_b3ba8f1b, phan2022stochastic} with a RBF kernel with kernel width $\sigma$. Obtained particles  $\tilde{X}_{kij}^{S}$ and $\tilde{X}_{ij}^{T}$ are then utilized to minimize the objective function in Eq. (\ref{eq:last_opt})
for updating $\psi=(\phi,\theta)$. Specifically, we utilize cross-entropy
for the classification loss term $\ell$ and the symmetric Kullback-Leibler
(KL) divergence for the local regularization term $s\left(X,\tilde{X};\psi\right)$
as
 $\frac{1}{2}KL\left(f_{\psi}\left(X\right)\Vert f_{\psi}\left(\tilde{X}\right)\right)+\frac{1}{2}KL\left(f_{\psi}\left(\tilde{X}\right)\Vert f_{\psi}\left(X\right)\right).$

Finally, the global-regularization function of interest $r^{g}\left(\left[X_{ki}^{S}\right]_{k,i},\left[X_{i}^{T}\right]_{i};\psi\right)$
is defined accordingly depending on the task and explicitly presented
in the sequel.

\subsection{Setting for Domain Adaptation and Semi-supervised Learning}

By considering the single source domain as the labeled portion and
the target domain as the unlabeled portion, the same setting can be
employed for DA and SSL. Particularly, we denote the data/label distribution
of the source domain or labeled portion by $\mathbb{P}_{1}^{S|l}$
and the data distribution of target domain or unlabeled portion by
$\mathbb{P}^{T|u}$. Notice that for SSL, $\mathbb{P}^{T|u}$ could
be the marginal of $\mathbb{P}^{S|l}$ by marginalizing out the label
dimension. Evidently, with this consideration, DA and SSL are special
cases of our general framework in Section \ref{subsec:Our-Framework},
where the global-regularization function of interest $r^{g}\left(\left[X_{i}^{S}\right]_{i},\left[X_{j}^{T}\right]_{j};\psi\right)$
is defined as 
\begin{equation}
\mathcal{W}_{d}\left(\frac{1}{B^{S}}\sum_{i=1}^{B^{S}}\delta_{U_{i}^{S}},\frac{1}{B^{T}}\sum_{j=1}^{B^{T}}\delta_{U_{j}^{T}}\right),\label{eq:global_DA_SSL}
\end{equation}
where $U_{i}^{S}=\left[g_{\phi}\left(X_{i}^{S}\right),h_{\theta}\left(g_{\phi}\left(X_{i}^{S}\right)\right)\right]$,
$U_{j}^{T}=\left[g_{\phi}\left(X_{j}^{T}\right),h_{\theta}\left(g_{\phi}\left(X_{j}^{T}\right)\right)\right]$,
and $\delta$ is the Dirac delta distribution. The cost metric $d$
is defined as 
\begin{align}
d\left(U_{i}^{S},U_{j}^{T}\right)  & :=\rho_{d}\left(g_{\phi}\left(X_{i}^{S}\right),g_{\phi}\left(X_{j}^{T}\right)\right)\nonumber
\\ & + \gamma\rho_{l}\left(h_{\theta}\left(g_{\phi}\left(X_{i}^{S}\right)\right),h_{\theta}\left(g_{\phi}\left(X_{j}^{T}\right)\right)\right), \label{eq:ot_cost}
\end{align}
where $\rho_{d}$ is a metric on the latent space and $\gamma>0$.

\begin{table*}[t]

\caption{Single domain generalization accuracy (\%) on CIFAR-10-C and CIFAR-100-C
datasets with different backbone architectures\label{tab:cifar-c-1}.
We use the\textbf{ bold} font to highlight the best results.}
\begin{center}
    \resizebox{1.9\columnwidth}{!}{
\begin{tabular}[width=1\textwidth]{c|cccccccccc}
\toprule
\multicolumn{1}{c}{Datasets} & Backbone & Standard & Cutout & CutMix & AutoDA & Mixup & AdvTrain & ADA & ME-ADA & GLOT-DR\tabularnewline
\midrule
\multirow{5}{*}{CIFAR-10-C} & AllConvNet & 69.2 & 67.1 & 68.7 & 70.8 & 75.4 & 71.9 & 73 & 78.2 & \textbf{82.5}\tabularnewline
 & DenseNet & 69.3 & 67.9 & 66.5 & 73.4 & 75.4 & 72.4 & 69.8 & 76.9 & \textbf{83.6}\tabularnewline
 & WideResNet & 73.1 & 73.2 & 72.9 & 76.1 & 77.7 & 73.8 & 79.7 & 83.3 & \textbf{84.4}\tabularnewline
 & ResNeXt & 72.5 & 71.1 & 70.5 & 75.8 & 77.4 & 73 & 78 & 83.4 & \textbf{84.5}\tabularnewline
\cmidrule{2-11} 
 & Average & 71 & 69.8 & 69.7 & 74 & 76.5 & 72.8 & 75.1 & 80.5 & \textbf{83.7}\tabularnewline
\midrule \midrule
\multirow{5}{*}{CIFAR-100-C} & AllConvNet & 43.6 & 43.2 & 44 & 44.9 & 46.6 & 44 & 45.3 & 51.2 & \textbf{54.8}\tabularnewline
 & DenseNet & 40.7 & 40.4 & 40.8 & 46.1 & 44.6 & 44.8 & 45.2 & 47.8 & \textbf{53.2}\tabularnewline
 & WideResNet & 46.7 & 46.5 & 47.1 & 50.4 & 49.6 & 44.9 & 50.4 & 52.8 & \textbf{56.5}\tabularnewline
 & ResNeXt & 46.6 & 45.4 & 45.9 & 48.7 & 48.6 & 45.6 & 53.4 & 57.3 & \textbf{58.4}\tabularnewline
\cmidrule{2-11} 
 & Average & 44.4 & 43.9 & 44.5 & 47.5 & 47.4 & 44.8 & 48.6 & 52.3 & \textbf{55.7}\tabularnewline
\bottomrule
\end{tabular}}
\end{center}

\end{table*}

With the global term in Eq. (\ref{eq:global_DA_SSL}), we aim to reduce
the discrepancy gap between the\emph{ source (labeled)} domain and
the \emph{target (unlabeled)} domain for learning domain-invariant
representations. It is worth noting that this global term in Eq. (\ref{eq:global_DA_SSL})
was inspected in DeepJDOT \citep{damodaran2018deepjdot} for DA setting.
Our approach is different from that approach in the local regularization term. 

\subsection{Setting for Domain Generalization}
By setting $B^{T}=0$ (i.e., not use any target data in training),
our general framework in Section \ref{subsec:Our-Framework} is applicable
to DG, wherein the global-regularization function of interest $r^{g}\left(\left[X_{ki}^{S}\right]_{k,i},\left[X_{i}^{T}\right]_{i};\psi\right)$
is
\begin{equation}
\sum_{m=1}^{M}\sum_{k=1}^{K}\frac{1}{K}\mathcal{W}_{d}\left(\tilde{\mathbb{P}}_{km},\tilde{\mathbb{P}}_{m}\right), \label{eq:global_DG}
\end{equation} 
where the cost metric $d=\rho_{d}$ is a metric on the latent space,
$\tilde{\mathbb{P}}_{km}$ is the empirical distribution over $g_{\phi}\left(X_{ki}^{S}\right)$
with $Y_{ki}^{S}=m$, and $\tilde{\mathbb{P}}_{m}=\frac{1}{K}\sum_{k=1}^{K}\tilde{\mathbb{P}}_{km}$.

\subsection{Setting for Adversarial Machine Learning}
For AML, we have only \emph{single source domain} and need to train
a deep model which is robust to adversarial examples. We denote the
data/label distribution of the source domain by $\mathbb{P}_{1}^{S}$
and propose using a dynamic and pseudo target domain of the \emph{on-the-fly
adversarial examples} $\left[\left[X_{1ij}^{S}\right]_{i=1}^{B^{S}}\right]_{j=1}^{n^{S}}$.
In addition to the local and loss terms as in Eq. (\ref{eq:last_opt}),
to strengthen model robustness, we propose the following global term
to move adversarial examples ($\sim\mathbb{P}^{T}$) to benign examples ($\sim\mathbb{P}_{1}^{S}$):
\begin{equation}
\label{eq:global_AML}
\mathcal{W}_{d}\left(\frac{1}{B_{1}^{S}}\sum_{i=1}^{B_{1}^{S}}\delta_{U_{i}^{S}},\frac{1}{B_{1}^{S}n^{S}}\sum_{i=1}^{B_{1}^{S}}\sum_{j=1}^{n^{S}}\delta_{U_{ij}^{S}}\right),
\end{equation}
where $U_{i}^{S}=\left[g_{\phi}\left(X_{1i}^{S}\right),h_{\theta}\left(g_{\phi}\left(X_{1i}^{S}\right)\right)\right]$,
$U_{ij}^{S}=\left[g_{\phi}\left(X_{1ij}^{S}\right),h_{\theta}\left(g_{\phi}\left(X_{1ij}^{S}\right)\right)\right]$,
and the metric $d$ is
\begin{multline}
d\left(U_{i}^{S},U_{\bar{i}j}^{S}\right)=\mathbb{I}_{Y_{1i}^{S}=Y_{1\bar{i}}^{S}}\Biggl[\rho_{d}\left(g_{\phi}\left(X_{1i}^{S}\right),g_{\phi}\left(X_{1\bar{i}j}^{S}\right)\right)\\
+\gamma\rho_{l}\left(h_{\theta}\left(g_{\phi}\left(X_{1i}^{S}\right)\right),h_{\theta}\left(g_{\phi}\left(X_{1\bar{i}j}^{S}\right)\right)\right)\Biggr],\label{eq:cost_AML}
\end{multline}
where $\mathbb{I}$ is the indicator function. Here we note that $X_{1\bar{i}j}^{S}$
is an adversarial example of $X_{1\bar{i}}^{S}$ which has the ground-truth
label $Y_{1\bar{i}}^{S}$, hence by using the cost metric as in Eq. 
(\ref{eq:cost_AML}), we encourage the adversarial example $X_{1\bar{i}j}^{S}$
to move to a group of the benign examples with the same label. 

Finally, to tackle the WS-related terms in equations. (\ref{eq:global_DA_SSL},\ref{eq:global_DG}, and \ref{eq:global_AML}), we employ the entropic
regularization dual form of WS, which was demonstrated to have favorable computational complexities~\citep{ Lin-2020-Revisiting, Lin-2019-Efficient,  Lin-2019-Efficiency}.

\section{Experiments}
To demonstrate the effectiveness of our proposed method, we evaluate
its performance on various experiment protocols, including DG, DA, SSL, and AML. Due to the space limitation, the detailed setup
regarding the architectures and hyperparameters are presented in the supplementary material\footnote{Our codes are available at \url{https://github.com/VietHoang1512/GLOT}}.  We tried to use the exact configuration of optimizers and hyper-parameters for all experiments and report the original results in prior work, if possible.

\begin{table*}[t]
\caption{Multi-source domain generalization accuracy (\%) on PACS datasets\label{tab:pacs-1}.
Each column title indicates the target domain used for evaluation,
while the rest are for training.}
\begin{center}
\resizebox{2\columnwidth}{!}{
\begin{tabular}{c|c|c|c|c|c|c|c|c|c|c|c|c}
\toprule
 & DSN & L-CNN & MLDG & Fusion & MetaReg & Epi-FCR & AGG & HEX & PAR & ADA & ME-ADA & GLOT-DR\tabularnewline
\midrule 
Art & 61.1 & 62.9 & 66.2 & 64.1 & 69.8 & 64.7 & 63.4 & 66.8 & 66.9 & 64.3 & \textbf{67.1} & 66.1 \tabularnewline
Cartoon & 66.5 & 67.0 & 66.9 & 66.8 & 70.4 & \textbf{72.3} & 66.1 & 69.7 & 67.1 & 69.8 & 69.9 & \textbf{72.3 }\tabularnewline
Photo & 83.3 & 89.5 & 88.0 & 90.2 & 91.1 & 86.1 & 88.5 & 87.9 & 88.6 & 85.1 & 88.6 & \textbf{90.4 }\tabularnewline
Sketch & 58.6 & 57.5 & 59.0 & 60.1 & 59.2 & 65.0 & 56.6 & 56.3 & 62.6 & 60.4 & 63.0 & \textbf{65.4 }\tabularnewline
\midrule 
Average & 67.4 & 69.2 & 70.0 & 70.3 & 72.6 & 72.0 & 68.7 & 70.2 & 71.3 & 69.9 & 72.2 & \textbf{73.5 }\tabularnewline
\bottomrule
\end{tabular}}

\end{center}

\end{table*}

\subsection{Experiments for DG}

In DG experiments, our setup closely follows \citep{zhao_maximum}.
In particular, we validate our method on the CIFAR-C single domain
generalization benchmark: train the model on either CIFAR-10 or CIFAR-100
dataset \citep{krizhevsky2009learning}, then evaluate it on CIFAR-10-C
or CIFAR-100-C \citep{hendrycks2019robustness}, correspondingly.
In terms of network architectures, we use the exact backbones from
\citep{zhao_maximum} to examine the versatility of our method that
can be adopted in any type of classifier. GLOT-DR is compared with other
state-of-the-art methods in image corruption robustness: Mixup \citep{zhang2018mixup},
Cutout \citep{devries2017improved} and Cutmix \citep{yun2019cutmix},
AutoDA \citep{cubuk2019autoaugment}, ADA \citep{volpi2018generalizing},
and ME-ADA \citep{zhao_maximum}.

Table \ref{tab:cifar-c-1} shows the average accuracy when we alternatively train the model on one category and evaluate on the rest. In every setting, GLOT-DR outperforms other methods by large margins. Specifically, our method exceeds the second-best method ME-ADA \citep{zhao_maximum} by $3.2\%$ on CIFAR-10-C and $3.4\%$ on CIFAR-100-C.  The substantial gain in terms of the accuracy on various backbone architectures demonstrates
the high applicability of our GLOT-DR.

Furthermore, we examine multi-source DG where the classifier needs
to generalize from multiple source domains to an unseen target domain on the PACS dataset \citep{li2017deeper}. Our proposed
method is applicable in this scenario since it is designed to better
learn domain invariant features as well as leverage the diversity
from generated data. We compare GLOT-DR against DSN \citep{bousmalis2016domain},
L-CNN \citep{li2017deeper}, MLDG \citep{li2018learning}, Fusion
\citep{mancini2018best}, MetaReg \citep{balaji2018metareg}, Epi-FCR,
AGG \citep{li2019episodic}, HEX \citep{wang2018learning}, and PAR
\citep{NEURIPS2019_3eefceb8}. Table \ref{tab:pacs-1} shows that
our GLOT-DR outperforms the baselines for three cases and averagely
surpasses the second-best baseline by $0.9\%$. The most noticeable
improvement is on the Sketch domain ($\thickapprox2.4\%)$, which
is the most challenging due to the fact that the styles of the images are colorless and
far different from the ones from Art Painting, Cartoon or Photos (i.e.,
larger domain shift).

\subsection{Experiments for DA}
In this section, we conduct experiments on the commonly used dataset
for real-world unsupervised DA - Office-31 \citep{saenko2010adapting},
comprising images from three domains: Amazon
(A), Webcam (W) and DSLR (D). Our proposed GLOT-DR is compared against
baselines: ResNet-50 \citep{he2016deep}, DAN \citep{pmlr-v37-long15},
RTN \citep{NIPS2016_ac627ab1}, DANN \citep{ganin2016domain}, JAN
\citep{long2017deep}, GTA \citep{sankaranarayanan2018generate},
CDAN \citep{long2017conditional}, DeepJDOT \citep{damodaran2018deepjdot}
and ETD \citep{li2020enhanced}. For a fair comparison, we follow
the training setups of CDAN and compare with other works using this
configuration. As can be seen from Table \ref{tab:office-31}, GLOT-DR
achieves the best overall performance among baselines with $87.8\%$
accuracy. Compared with ETD, which is another OT-based domain adaptation
method, our performance significantly increase by $4.1\%$ on A\textrightarrow W
task, $2.1\%$ on W\textrightarrow A and $1.6\%$ on average.

\begin{table}[!ht]
   \vspace{-5mm}    
    \centering
        \caption{Accuracy (\%) on Office-31 \citep{saenko2010adapting} of ResNet50
model \citep{he2016deep} in unsupervised DA methods.}
    \resizebox{1\columnwidth}{!}{
    \begin{tabular}{c|cccccc|c}
    \toprule
    Method & A\textrightarrow W & D\textrightarrow W & W\textrightarrow D & A\textrightarrow D & D\textrightarrow A & W\textrightarrow A & Avg\tabularnewline
    \midrule
    ResNet & 68.4 & 96.7 & 99.3 & 68.9 & 62.5 & 60.7 & 76.1\tabularnewline
    DAN & 80.5 & 97.1 & 99.6 & 78.6 & 63.6 & 62.8 & 80.4\tabularnewline
    RTN & 70.2 & 96.6 & 95.5 & 66.3 & 54.9 & 53.1 & 72.8\tabularnewline
    DANN & 84.5 & 96.8 & 99.4 & 77.5 & 66.2 & 64.8 & 81.6\tabularnewline
    JAN & 82 & 96.9 & 99.1 & 79.7 & 68.2 & 67.4 & 82.2\tabularnewline
    GTA & 89.5 & 97.9 & 99.8 & 87.7 & 72.8 & \textbf{71.4} & 86.5\tabularnewline
    CDAN & 93.1 & 98.2 & \textbf{100} & \textbf{89.8} & 70.1 & 68 & 86.6\tabularnewline
    DeepJDOT & 88.9 & 98.5 & 99.6 & 88.2 & \textbf{72.1} & 70.1 & 86.2\tabularnewline
    ETD & 92.1 & \textbf{100} & \textbf{100} & 88 & 71 & 67.8 & 86.2\tabularnewline
    \midrule 
    GLOT-DR & \textbf{96.2 } & 98.9  & \textbf{100 } & 90.6  & 69.9  & 69.6  & \textbf{87.8 }\tabularnewline
    \bottomrule
\end{tabular}}
    \vspace{-5mm} 
    \label{tab:office-31}
\end{table}

\begin{figure}[!ht]
    \centering
     \includegraphics[width=0.85\columnwidth]{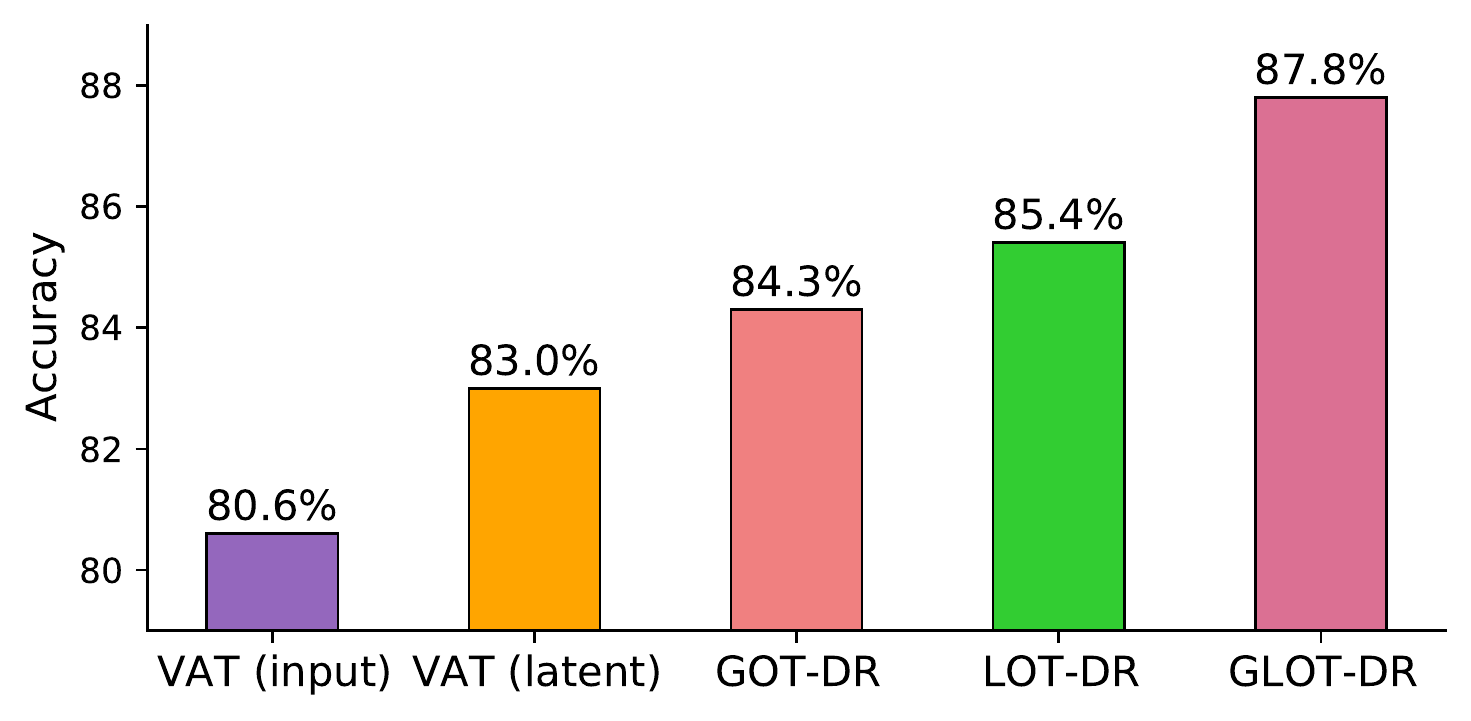}
    \caption{Average accuracy of ResNet50 \citep{he2016deep} on Office-31: Comparision
between GLOT-DR's variants and VAT \citep{VAT} on the input and latent spaces.}
    \label{fig:office-31-ablation}
    \vspace{-3mm}

\end{figure}

We further extensively investigate the role of different components
in GLOT-DR. Specifically, the elimination of the global-regularization
term in equation (\ref{eq:global_DA_SSL}) downgrades our method to Local Optimal
Transport based Distributional Robustness (LOT-DR). Similarly, when
discarding the local distribution robustness term, the attained method
is denoted by GOT-DR. We then compare these 2 variants of GLOT-DR
to the well-known adversarial machine learning method VAT \citep{VAT}.
To be more specific, in the adversarial samples generation, we apply
VAT by perturbing on the: (i) input space, (ii) latent space. Figure
\ref{fig:office-31-ablation} shows that the employment of VAT on
latent space (orange) is more effective than on the input space (purple),
$83\%$ and $80.6\%$. However, using GOT-DR or LOT-DR is even more
effective: performance is boosted to $84.3\%$ and $85.4\%$, respectively.
Lastly, using the full method GLOT-DR yields the highest average accuracy
score among all. 

\subsection{Experiments for SSL}
\label{sec:ssl}
\begin{figure}[!ht]
\centering{}\includegraphics[width=0.95\columnwidth]{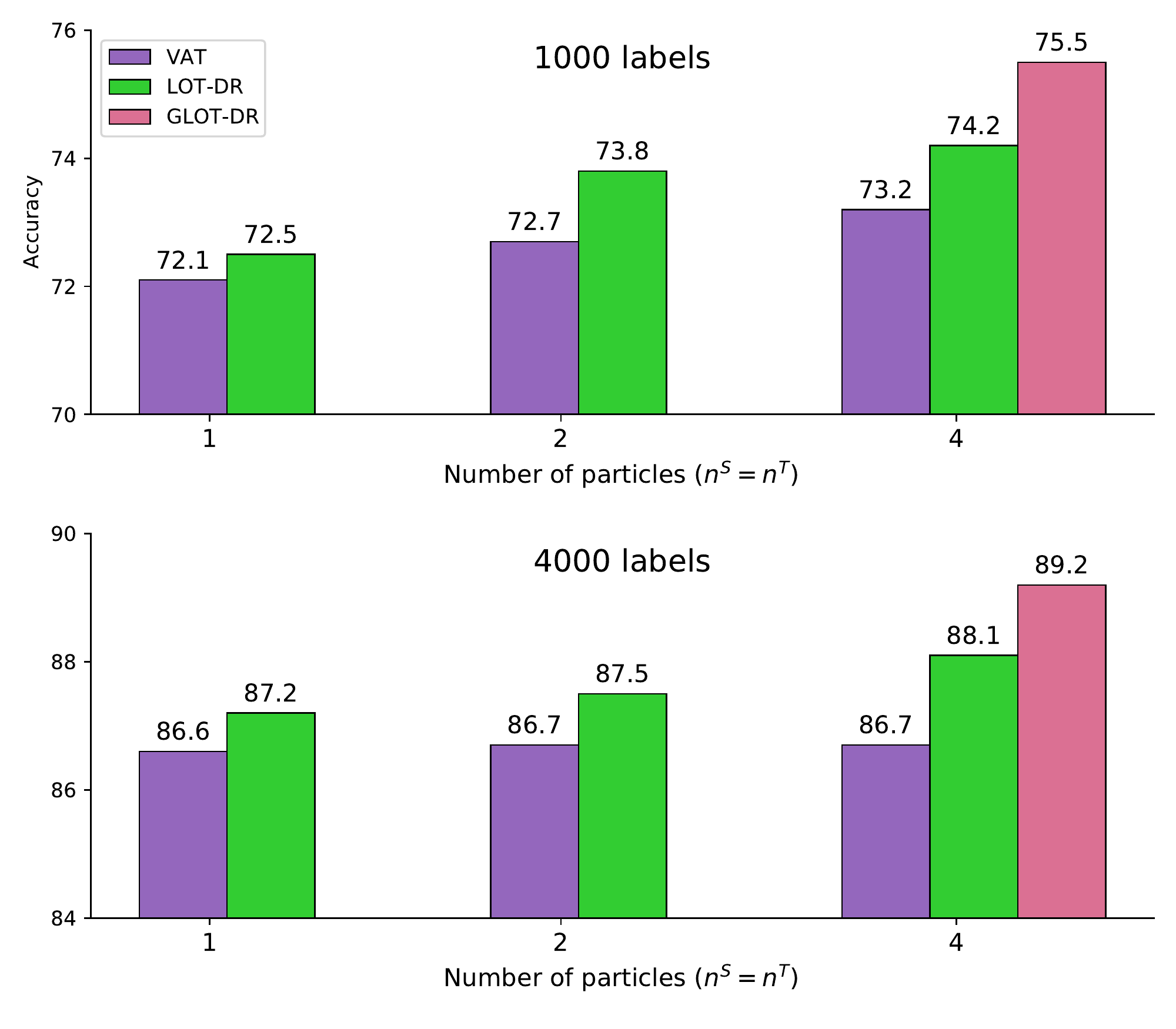}
\caption{Accuracy (\%) on CIFAR-10 of ConvLarge model in SSL settings when
using 1,000 and 4,000 labeled examples (i.e. 100 and 400 labeled samples
each class). Best viewed in color. \label{fig:ssl}}
\vspace*{-8mm}  
\end{figure}

Sharing a similar objective with DA, which utilizes the unlabeled
samples for improving the model performance, SSL methods can also
benefit from our proposed technique. We present
the empirical results on CIFAR-10 benchmark with ConvLarge architecture,
following VAT's protocol \citep{VAT}, which serves as a strong baseline
in this experiment. We refer readers to the supplementary material
for more details on the architecture of ConvLarge. Results in Figure
\ref{fig:ssl} (when training with 1,000 and 4,000 labeled examples)
demonstrate that, with only $n^{S}=n^{T}=1$ perturbed sample per
anchor, the performance of LOT-DR slightly outperforms VAT with $\sim0.5\%$.
With more perturbed samples per anchor, this gap increases: approximately
$1\%$ when $n^{S}=n^{T}=2$ and $1.5\%$ when $n^{S}=n^{T}=4$. Similar
to the previous DA experiment, adding the global regularization term
helps increase accuracy by $\sim1\%$ in this setup.

\subsection{Experiments for AML}
Table \ref{tab:aml-cf10-r18} shows the evaluation against adversarial
examples. 

\noindent We compare our method with PGD-AT \citep{madry2017towards}
and TRADES \citep{zhang2019theoretically}, two well-known defense methods in AML and SAT  \citep{bouniot2021optimal}. For the sake of fair comparison, we use the same adversarial training setting for all methods, which is carefully
investigated in \citep{pang2020bag}.  We also compare with
adversarial distributional training methods \citep{deng2020adversarial}
(ADT-EXP and ADT-EXPAM), which assume that the adversarial distribution
explicitly follows normal distribution. It can be seen from Table
\ref{tab:aml-cf10-r18} that our GLOT-DR method outperforms all these
baselines in both natural and robustness performance.  Specifically,
compared to PGD-AT, our method has an improvement of 0.8\% in natural accuracy and around 1\% robust accuracies against PGD200 and AA attacks. Compared to TRADES, while achieving the same level of robustness, our method has a better performance with benign examples with a gap of 2.5\%. Especially, our method significantly outperforms ADT by around 7\% under the PGD200 attack.

\begin{table}[!ht]
\centering
\caption{Adversarial robustness evaluation on CIFAR10 of ResNet18 model. PGD,
AA and B\&B represent the robust accuracy against the PGD attack (with
10/200 iterations) \citep{madry2017towards}, Auto-Attack \citep{croce2020reliable}
and B\&B attack \citep{brendel2019accurate}, respectively, while
NAT denotes the natural accuracy. Note that $^{\star}$ results are taken from Pang et al. \citep{pang2020bag}, while $^{\diamond}$ results are our reproduced results.
 \label{tab:aml-cf10-r18}}
 \vspace*{-1mm} 
\resizebox{1\columnwidth}{!}{
\begin{tabular}{c|ccccc}
\toprule
Method & NAT & PGD10 & PGD200 & AA & B\&B\tabularnewline
\midrule
$\text{PGD-AT}^{\star}$ & 82.52 & 53.58 & - & 48.51 & -\tabularnewline
$\text{TRADES}^{\star}$ & 81.45 & 53.51 & - & 49.06 & -\tabularnewline
$\text{PGD-AT}^{\diamond}$ & 83.36 & 53.52 & 52.21 & 49.00 & 48.50\tabularnewline
$\text{TRADES}^{\diamond}$ & 81.64 & 53.73 & 53.11 & 49.77 & 49.02\tabularnewline
$\text{ADT-EXP}$ & 83.02 & - & 45.80 & 45.80 & 46.50\tabularnewline
$\text{ADT-EXPAM}$ & 84.11 & - & 46.10 & 44.50 & 45.83\tabularnewline
SAT & 83.45 & 53.95 & 51.37 & 48.80 & \textbf{49.40}  \tabularnewline
\midrule 
GLOT-DR & \textbf{84.13} & \textbf{54.13} & \textbf{53.1}8 & \textbf{49.94} & \textbf{49.40}\tabularnewline
\bottomrule
\end{tabular}}
  \vspace*{-3mm}  
\end{table}


\section{Conclusion}
Although DR is a promising framework to improve neural network robustness and generalization capability, its current formulation shows some limitations, circumventing its application to real-world problems. Firstly, its formulation is not sufficiently rich to express a global regularization effect targeting many applications. Secondly, the dual form is not readily trainable to incorporate into the training of deep learning models. In this work, we propose a rich OT based DR framework, named
\textbf{\emph{G}}\emph{lobal-}\textbf{\emph{L}}\emph{ocal }\textbf{\emph{O}}\emph{ptimal
}\textbf{\emph{T}}\emph{ransport based }\textbf{\emph{D}}\emph{istributional
}\textbf{\emph{R}}\emph{obustness} (GLOT-DR) which is sufficiently
rich for many real-world applications including DG, DA, SSL, and AML
and has a closed-form solution. Finally, we conduct comprehensive experiments to compare our GLOT-DR with state-of-the-art baselines accordingly. Empirical results have demonstrated the merits of our GLOT-DR  on standard benchmark datasets .

\newpage
\paragraph{Acknowledgements.} Trung Le and Dinh Phung were supported by the US Air Force grant FA2386-21-1-4049. Trung Le was supported by the ECR Seed grant of the Faculty of Information Technology, Monash University. Trung Le and Dinh Phung were also supported by the Australian Research Council (ARC) DP230101176 grant.

\bibliography{ref}

\newpage
\appendix
\onecolumn

\begin{center}
{\bf \large Supplement to ``Global-Local Regularization Via Distributional Robustness''}
\end{center}

These appendices provide supplementary details and results of GLOT, including our theory development and additional experiments. This consists of the following sections:
\begin{itemize}
\item Appendix \ref{sec:Proofs} contains the proofs of our theory development.
\item Appendix \ref{sec:Additional-Experiments} contains the network architectures,
experiment settings of our experiments and additional ablation studies.
\end{itemize}

\section{Proofs of Our Theory Development \label{sec:Proofs}}
We here give the proof for the equivalence in optimizing two equations (\ref{eq:dro_p1}) and (\ref{eq:dro_p2}) in Section \ref{subsec:dro_p12}.  Then, we detail how to derive the optimization formulations (\ref{thm:opt_sol}) and (\ref{eq:last_opt})  for
solving the problems discussed in Section \ref{subsec:Our-Framework}.
\subsection{Proof of Lemma \ref{lem:dro_p2} \label{subsec:dro_p12}}

Let 
\[
\gamma^{*}=\argmax{\gamma:\in\Gamma_{\epsilon}}\underset{\left(Z,\tilde{Z}\right)\sim\gamma}{\mathbb{E}}\left[r\left(\tilde{Z};\phi,\theta\right)\right]
\]
be the optimal solution of the inner max in equation (\ref{eq:dro_p2}). Denote
$\tilde{\mathbb{P}}^{*}$ as the distribution obtained from $\gamma^{*}$
by maginalizing the dimensions of $Z$. We prove that $\tilde{\mathbb{P}}^{*}$
is the optimal solution of the inner max in equation (\ref{eq:dro_p1}). Let
$\tilde{\mathbb{P}}$ be a feasible solution of the inner max in equation (\ref{eq:dro_p1}),
meaning that $\mathcal{W}_{\rho}\left(\mathbb{P},\tilde{\mathbb{P}}\right)\leq\epsilon$.
Therefore, there exists $\gamma\in\Gamma$$\left(\mathbb{P},\tilde{\mathbb{P}}\right)$
such that $\underset{\left(Z,\tilde{Z}\right)\sim\gamma}{\mathbb{E}}\left[\rho\left(Z,\tilde{Z}\right)\right]^{1/q}\leq\epsilon$
or $\gamma\in\Gamma_{\epsilon}$. We have
\begin{align*}
\underset{\tilde{\mathbb{P}}}{\mathbb{E}}\left[r\left(\tilde{Z};\phi,\theta\right)\right] & =\underset{\gamma}{\mathbb{E}}\left[r\left(\tilde{Z};\phi,\theta\right)\right] \leq  \underset{\gamma^{*}}{\mathbb{E}}\left[r\left(\tilde{Z};\phi,\theta\right)\right]=\underset{\mathbb{P}^{*}}{\mathbb{E}}\left[r\left(\tilde{Z};\phi,\theta\right)\right].
\end{align*}

We reach the conclusion that $\tilde{\mathbb{P}}^{*}$ is the optimal
solution of the inner max in equation (\ref{eq:dro_p1}). That concludes our
proof.

\subsection{Proof of Theorem \ref{thm:opt_sol}}

Given $\gamma\in\Gamma_{\epsilon}$, we first prove that if $\underset{\left(Z,\tilde{Z}\right)\sim\gamma}{\mathbb{E}}\left[\rho\left(Z,\tilde{Z}\right)\right]$
is finite $\forall q>1$ then
\begin{align*}
M_{\gamma} & :=\lim_{q\goto\infty}\underset{\left(Z,\tilde{Z}\right)\sim\gamma}{\mathbb{E}}\left[\rho\left(Z,\tilde{Z}\right)\right]^{1/q}  = \sup_{\left(Z,\tilde{Z}\right)\in\support\left(\gamma\right)}\max\left\{ \max_{k,i,j}\norm{X_{kij}^{S}-\tilde{X}_{kij}^{S}}_{p},\max_{i,j}\norm{X_{ij}^{T}-\tilde{X}_{ij}^{T}}_{p}\right\}.
\end{align*}
Let denote $A_{\gamma}$ as the set of $\left(Z,\tilde{Z}\right)\in\support\left(\gamma\right)$
such that 
\[
\max\left\{ \max_{k,i,j}\norm{X_{kij}^{S}-\tilde{X}_{kij}^{S}}_{p},\max_{i,j}\norm{X_{ij}^{T}-\tilde{X}_{ij}^{T}}_{p}\right\} =M_{\gamma}.
\]

We have 
\begin{align*}
 & \underset{\left(Z,\tilde{Z}\right)\sim\gamma}{\mathbb{E}}\left[\rho\left(Z,\tilde{Z}\right)\right]^{1/q}
= \left[\int_{A_{\gamma}}\rho\left(Z,\tilde{Z}\right)d\gamma\left(Z,\tilde{Z}\right)+\int_{A_{\gamma}^{c}}\rho\left(Z,\tilde{Z}\right)d\gamma\left(Z,\tilde{Z}\right)\right]^{1/q}.
\end{align*}
\newpage
It is obvious that if $\left(Z,\tilde{Z}\right)\sim\gamma$ then
\begin{alignat*}{1}
\rho\left(Z,\tilde{Z}\right) & :=\sum_{i=1}^{B^{T}}\sum_{j=1}^{n^{T}}\norm{X_{ij}^{T}-\tilde{X}_{ij}^{T}}_{p}^{q}
 +\sum_{k=1}^{K}\sum_{i=1}^{B_{k}^{S}}\sum_{j=1}^{n^{S}}\norm{X_{kij}^{S}-\tilde{X}_{kij}^{S}}_{p}^{q}.
\end{alignat*}

Therefore, for $\left(Z,\tilde{Z}\right)\in A_{\gamma}^{c}$, we have
\[
\lim_{q\goto\infty}\frac{\rho\left(Z,\tilde{Z}\right)}{M_{\gamma}^{q}}=0,
\]
while for $\left(Z,\tilde{Z}\right)\in A_{\gamma}$, we have
\[
\lim_{q\goto\infty}\frac{\rho\left(Z,\tilde{Z}\right)}{M_{\gamma}^{q}}=1.
\]

We derive as
\begin{align*}
 \lim_{q\goto\infty}\underset{\left(Z,\tilde{Z}\right)\sim\gamma}{\mathbb{E}}\left[\rho\left(Z,\tilde{Z}\right)\right]^{1/q}
& =  M_{\gamma}\lim_{q\goto\infty}\left[\int_{A_{\gamma}}\frac{\rho\left(Z,\tilde{Z}\right)}{M^{q}}d\gamma\left(Z,\tilde{Z}\right)+\int_{A_{\gamma}^{c}}\frac{\rho\left(Z,\tilde{Z}\right)}{M^{q}}d\gamma\left(Z,\tilde{Z}\right)\right]^{1/q}\\
& =  M_{\gamma}\lim_{q\goto\infty}\gamma\left(A_{\gamma}\right)^{1/q}=M_{\gamma}.
\end{align*}

Therefore, $\gamma\in\Gamma_{\epsilon}$ with $q=\infty$ is equivalent
to the fact that the support set $\support\left(\gamma\right)$ is
the union of $B_{Z}$ with $Z\in\support\left(\mathbb{P}\right)$,
where $B_{Z}$ is defined as follows:
\begin{align*}
B_{Z} & :=\prod_{k=1}^{K}\prod_{i=1}^{B_{k}^{S}}\prod_{j=0}^{n_{k}^{S}}B_{\epsilon}\left(X_{kij}^{S}\right)\prod_{i=1}^{B^{T}}\prod_{j=1}^{n^{T}}B_{\epsilon}\left(X_{ij}^{T}\right) =\prod_{k=1}^{K}\prod_{i=1}^{B_{k}^{S}}\prod_{j=0}^{n_{k}^{S}}B_{\epsilon}\left(X_{ki}^{S}\right)\prod_{i=1}^{B^{T}}\prod_{j=1}^{n^{T}}B_{\epsilon}\left(X_{i}^{T}\right).
\end{align*}

We can equivalently turn the optimization problem in equation (\ref{eq:dro_entropic})
as follows:
\begin{align}
 \hspace{2.5cm}\max_{\gamma\in\Gamma}\underset{\left(Z,\tilde{Z}\right)\sim\gamma}{\mathbb{E}}\left[r\left(\tilde{Z};\phi,\theta\right)\right]+\frac{1}{\lambda}\mathbb{H}\left(\gamma\right)
\hspace{1.5cm}\text{s.t.}:  \support\left(\gamma\right)=\underset{Z\in \: \support\left(\mathbb{P}\right)}{\bigcup}B_{Z}.\label{eq:equi_entropic}
\end{align}
where $\Gamma=\cup_{\tilde{\mathbb{P}}}\Gamma\left(\mathbb{P},\tilde{\mathbb{P}}\right)$.

Because $\gamma\in\Gamma\left(\mathbb{P},\tilde{\mathbb{P}}\right)$
for some $\tilde{\mathbb{P}}$, we can parameterize its density function
as:
\[
\gamma\left(Z,\tilde{Z}\right)=p\left(Z\right)\tilde{p}\left(\tilde{Z}\mid Z\right),
\]
where $p\left(Z\right)$ is the density function of $\mathbb{P}$
and $\tilde{p}\left(\tilde{Z}\mid Z\right)$ has the support set $B_{Z}$.
Please note that the constraint for $\tilde{p}\left(\tilde{Z}\mid Z\right)$
is $\int_{B_{Z}}\tilde{p}\left(\tilde{Z}\mid Z\right)d\tilde{Z}=1$.

The Lagrange function for the optimization problem in equation (\ref{eq:equi_entropic})
is as follows:
\begin{align*}
\mathcal{L} & =\int r\left(\tilde{Z};\phi,\theta\right)p\left(Z\right)\tilde{p}\left(\tilde{Z}|Z\right)dZd\tilde{Z} -\frac{1}{\lambda}\int p\left(Z\right)\tilde{p}\left(\tilde{Z}|Z\right)\log\left[p\left(Z\right)\tilde{p}\left(\tilde{Z}|Z\right)\right]dZd\tilde{Z}\\
& \hspace*{56.5mm} + \int\alpha\left(Z\right)\left[\tilde{p}\left(\tilde{Z}\mid Z\right)d\tilde{Z}-1\right]d\tilde{Z}dZ,
\end{align*}
where the integral w.r.t $Z$ over on $\support\left(\mathbb{P}\right)$
and the one w.r.t. $\tilde{Z}$ over $B_{Z}$. 

Taking the derivative of $\mathcal{L}$ w.r.t. $\tilde{p}\left(\tilde{Z}\mid Z\right)$
and setting it to $0$, we obtain
\begin{align*}
0 & =r\left(\tilde{Z};\phi,\theta\right)p\left(Z\right)+\alpha\left(Z\right)
  -\frac{p\left(Z\right)}{\lambda}\left[\log p\left(Z\right)+\log\tilde{p}\left(\tilde{Z}|Z\right)+1\right], \\
\tilde{p}\left(\tilde{Z}|Z\right) & =\frac{\exp\left\{ \lambda\left[r\left(\tilde{Z};\phi,\theta\right)+\frac{\alpha\left(Z\right)}{p\left(Z\right)}\right]-1\right\} }{p\left(Z\right)}.
\end{align*}
Taking into account $\int_{B_{Z}}\tilde{p}\left(\tilde{Z}\mid Z\right)d\tilde{Z}=1$,
we achieve
\[
\int_{B_{Z}}\exp\left\{ \lambda r\left(\tilde{Z};\phi,\theta\right)\right\} d\tilde{Z}=\frac{p\left(Z\right)}{\exp\left\{ \lambda\frac{\alpha\left(Z\right)}{p\left(Z\right)}-1\right\} }.
\]

Therefore, we arrive at
\[
\tilde{p}^{*}\left(\tilde{Z}|Z\right)=\frac{\exp\left\{ \lambda r\left(\tilde{Z};\phi,\theta\right)\right\} }{\int_{B_{Z}}\exp\left\{ \lambda r\left(\tilde{Z};\phi,\theta\right)\right\} d\tilde{Z}},
\]
\begin{equation}
\gamma^{*}\left(Z,\tilde{Z}\right)=p\left(Z\right)\frac{\exp\left\{ \lambda r\left(\tilde{Z};\phi,\theta\right)\right\} }{\int_{B_{Z}}\exp\left\{ \lambda r\left(\tilde{Z};\phi,\theta\right)\right\} d\tilde{Z}}.\label{eq:optimal_gamma}
\end{equation}

Finally, by noting that

\[
p\left(Z\right)=\prod_{k=1}^{K}\prod_{i=1}^{B_{k}^{S}}\prod_{j=0}^{n_{k}^{S}}p_{k}^{S}\left(X_{ki}^{S},Y_{ki}^{S}\right)\prod_{i=1}^{B^{T}}\prod_{j=0}^{n^{T}}p^{T}\left(X_{i}^{T}\right)\exp\left\{ \lambda r\left(\tilde{Z};\phi,\theta\right)\right\}
\]
\begin{align*}
  =\exp\left\{ \lambda\beta r^{g}\left(\tilde{Z};\psi\right)\right\}  \prod_{k=1}^{K}\prod_{i=1}^{B_{k}^{S}}\prod_{j=0}^{n_{k}^{S}}\exp\left\{ \lambda[\alpha s(X_{ki}^{S},\tilde{X}_{kij}^{S};\psi)+\ell(\tilde{X}_{kij}^{S},Y_{ki}^{S};\psi)]\right\} \prod_{i=1}^{B^{T}}\prod_{j=1}^{n^{T}}\exp\left\{ \lambda\alpha s\left(X_{i}^{T},\tilde{X}_{ij}^{T};\psi\right)\right\} .
\end{align*}

And

\begin{align*}
 \int_{B_{Z}}\exp\left\{ \lambda r\left(\tilde{Z};\phi,\theta\right)\right\} d\tilde{Z}=\exp\left\{ \lambda\beta r^{g}\left(\tilde{Z};\psi\right)\right\} \prod_{k=1}^{K}\prod_{i=1}^{B_{k}^{S}}\prod_{j=0}^{n_{k}^{S}}
 \int_{B_{\epsilon}\left(X_{ki}^{S}\right)}\exp\left\{ \lambda[\alpha s(X_{ki}^{S},\tilde{X}_{kij}^{S};\psi)+\ell(\tilde{X}_{kij}^{S},Y_{ki}^{S};\psi)]\right\} d\tilde{X}_{kij}^{S}\\
 \prod_{i=1}^{B^{T}}\prod_{j=1}^{n^{T}}\int_{B_{\epsilon}\left(X_{i}^{T}\right)}\exp\left\{ \lambda\alpha s\left(X_{i}^{T},\tilde{X}_{ij}^{T};\psi\right)\right\} d\tilde{X}_{ij}^{T},
\end{align*}
we reach the conclusion.

\subsection{Proof of the optimization problem in equation~(\ref{eq:last_opt})}

By substituting $\gamma^{*}\left(Z,\tilde{Z}\right)$ in equation (\ref{eq:optimal_gamma})
back to the objective function in (\ref{eq:dro_p2}), we obtain\\
\[
\min_{\psi}\,\min_{\theta,\phi}\max_{\gamma:\in\Gamma_{\epsilon}}\underset{\left(Z,\tilde{Z}\right)\sim\gamma^{*}}{\mathbb{E}}\left[r\left(\tilde{Z};\phi,\theta\right)\right].
\]

By referring to the construction of $Z$ and $\tilde{Z}$ and noting
that for $\left(Z,\tilde{Z}\right)\sim\gamma^{*}$
\begin{align*}
r^{l}\left(\tilde{Z};\psi\right)  & :=\sum_{i=1}^{B^{T}}\sum_{j=1}^{n^{T}}s\left(\tilde{X}_{i0}^{T},\tilde{X}_{ij}^{T};\psi\right)
+ \sum_{k=1}^{K}\sum_{i=1}^{B_{k}^{S}}\sum_{j=1}^{n_{k}^{S}}s\left(\tilde{X}_{ki0}^{S},\tilde{X}_{kij}^{S};\psi\right)\\
& = \sum_{i=1}^{B^{T}}\sum_{j=1}^{n^{T}}s\left(X_{i}^{T},\tilde{X}_{ij}^{T};\psi\right)
+ \sum_{k=1}^{K}\sum_{i=1}^{B_{k}^{S}}\sum_{j=1}^{n_{k}^{S}}s\left(X_{ki}^{S},\tilde{X}_{kij}^{S};\psi\right), \\
\mathcal{L}\left(\tilde{Z};\psi\right) & :=\sum_{k=1}^{K}\sum_{i=1}^{B_{k}^{S}}\sum_{j=0}^{n_{k}^{S}}\ell\left(\tilde{X}_{kij}^{S},Y_{ki}^{S};\psi\right).
\end{align*}
As a consequence, we gain the final optimization problem.

\section{Implementation Details \label{sec:Additional-Experiments}}
In this section, we provide the detailed implementation for all of our experiments along with some additional experimental results. We begin with presenting the pseudo code used to sample from local distributions of our method.

\begin{algorithm}[H]
\caption{Projected SVGD.\label{alg:svgd} }
\begin{algorithmic} 

\STATE \textbf{Input:} A local distribution around $X$ with an unnormalized
density function $\tilde{p}(\cdot)$ and a set of initial particles
$\{X_{i}^{0}\}_{i=1}^{n}$.

\STATE \textbf{Output:} A set of particles $\{X_{i}\}_{i=1}^{n}$
that approximates the local distribution corresponding to $\tilde{p}(\cdot)$.

\FOR{$l=1$ to $L$}

\STATE $X_{i}^{l+1}=\prod_{B_{\epsilon}\left(X\right)}\left[X_{i}^{l}+\eta_{l}\hat{\phi}^{*}(X_{i}^{l})\right]$ 

where $\hat{\phi}^{*}(X)=\frac{1}{n}\sum_{j=1}^{n}[k(X_{j}^{l},X)\triangledown_{X_{j}^{l}}\log\tilde{p}(X_{j}^{l})+\triangledown_{X_{j}^{l}}k(X_{j}^{l},X)]$
and $\eta_{l}$ is the step size at the $l^{\text{th}}$ iteration.

\ENDFOR

\end{algorithmic} 
\end{algorithm}

\subsection{Entropic Regularized Duality for WS}

To enable the application of optimal transport in machine learning
and deep learning, Genevay et al. developed an entropic regularized
dual form in \citep{stochastic_ws}. First, they proposed to add an
entropic regularization term to the primal form: 
\begin{equation}
\mathcal{W}_{d}^{\epsilon}\left(\mathbb{P},\mathbb{Q}\right)  :=\min_{\gamma\in\Gamma\left(\mathbb{Q},\mathbb{P}\right)}\Biggl\{\underset{\left(\bx,\by\right)\sim\gamma}{\mathbb{E}}\left[d\left(\bx,\by\right)\right]\nonumber+\epsilon D_{KL}\left(\gamma\Vert\mathbb{P}\otimes\mathbb{Q}\right)\Biggr\} 
\label{eq:entropic_primal}
\end{equation}
where $\epsilon$ is the regularization rate, $D_{KL}\left(\cdot\Vert\cdot\right)$
is the Kullback-Leibler (KL) divergence, and $\mathbb{P}\otimes\mathbb{Q}$
represents the specific coupling in which $\mathbb{Q}$ and $\mathbb{P}$
are independent. Note that when $\epsilon\goto0$, $\mathcal{W}_{d}^{\epsilon}\left(\mathbb{P},\mathbb{Q}\right)$
approaches $\mathcal{W}_{d}\left(\mathbb{P},\mathbb{Q}\right)$ and
the optimal transport plan $\gamma_{\epsilon}^{*}$ of equation (\ref{eq:entropic_primal})
also weakly converges to the optimal transport plan $\gamma^{*}$
of the primal form. In practice, we set $\epsilon$ to be a small
positive number, hence $\gamma_{\epsilon}^{*}$ is very close to $\gamma^{*}$.
Second, using the Fenchel-Rockafellar theorem, they obtained the following
dual form w.r.t. the potential $\phi$
\vspace{1cm}
\begin{align}
\mathcal{W}_{d}^{\epsilon}\left(\mathbb{P},\mathbb{Q}\right) & =\max_{\phi}\left\{ \int\phi_{\epsilon}^{c}\left(\bx\right)\mathrm{d}\mathbb{Q}\left(\bx\right)+\int\phi\left(\by\right)\mathrm{d}\mathbb{P}\left(\by\right)\right\} \nonumber \\
 & =\max_{\phi}\left\{ \underset{\mathbb{Q}}{\mathbb{E}}\left[\phi_{\epsilon}^{c}\left(\bx\right)\right]+\underset{\mathbb{P}}{\mathbb{E}}\left[\phi\left(\by\right)\right]\right\} ,\label{eq:entropic_dual}
\end{align}
where $\phi_{\epsilon}^{c}\left(\bx\right):=-\epsilon\log\left(\underset{{\mathbb{P}}}{\mathbb{E}}\left[\exp\left\{ \frac{-d\left(\bx,\by\right)+\phi\left(\by\right)}{\epsilon}\right\} \right]\right)$.

In order to calculate the global WS-related regularization terms in equations. . (11, 13, and 14),
we apply the above entropic regularized dual form. The Kantorovich
potential network $\phi$ is a simple network with two fully connected
layers with ReLU activation in the middle: $\mathrm{FC_{latent\_dim\times512}}\rightarrow\mathrm{ReLU}\rightarrow\mathrm{FC_{512\times1}}$
is used throughout experiments. Note that the $\mathrm{latent\_dim}$
depends on the main network.

Additionally, the distance $\rho_{d}$ in equation (\ref{eq:ot_cost})
used in all experiments is the Euclidean distance $\mathrm{d(x_{1},x_{2})=||x_{1}-x_{2}||_{2}^{2}}$
, the prediction discrepancy trade-off $\gamma$ is set equal to $0.5$,
and the entropic regularization parameter $\lambda$ in equation (\ref{eq:dro_entropic})
is $0.1$.

\subsection{Projected SVGD Setting}

For Projected SVGD in Algorithm \ref{alg:svgd}, we employ an RBF
kernel
\[
k\left(X,\tilde{X}\right)=\exp\left\{ \frac{-\norm{X-\tilde{X}}_{2}^{2}}{2\sigma^{2}}\right\} ,
\]
where the kernel width is set according to the main paper \citep{NIPS2016_b3ba8f1b}.

\subsection{Experiments for DG}

\subsubsection{Network Architecture and Hyperparamters}

As mentioned in the main paper, we incorporate well-studied backbones
for our experiments, following the implementation of for single domain
generalization tasks in \citep{zhao_maximum}. In particular:
\begin{itemize}
\item LeNet5 \citep{lecun1989backpropagation} is employed in the MNIST
experiment. We first pre-train the network on the MNIST dataset without
applying any DG method for 100 iterations, then on each iteration
100, 200, 300 we generate particles with $n^{S}=n^{T}=n\in\left\{ 1,2,4\right\} $
by running the Projected SVGD sampling\ref{alg:svgd} in $L=15$ iterations,
step size $\eta=0.002$. We use Adam optimizer \citep{kingma2014adam}
with learning rate $10^{-5}$ and train for 15000 iteration in total
with batch size of $32$. 
\item CIFAR-C \footnote{Note that in both CIFAR-C and MNIST experiments, we are provided with
only a single source domain, thus GLOT-DR downgrades exactly to LOT-DR.} experiment uses 4 different backbone architectures, namely: All Convolutional
Network (AllConvNet) \citep{springenberg2014striving}, DenseNet \citep{huang2017densely}
, WideResNet \citep{zagoruyko2016wide}, and ResNeXt \citep{xie2017aggregated}.
We set $n^{S}=n^{T}=n=2$ particles, $L=15$ iterations, step size
$\eta=0.001$ and minimize the loss with SGD optimizer with initial
learning rate of $0.1$ and batch size 128. Similar to MNIST experiment,
we first pretrain the network for 10 epochs then generate augmented
images on epoch 10 and 20, number of total epochs required for training
are 150 in the case of AllConvNet and WideResNet, 250 epochs for DenseNet
and ResNeXt. 
\item We used an AlexNet \citep{krizhevsky2012imagenet} pretrained on ImageNet
\citep{russakovsky2015imagenet} in the PACS experiment. Different
from the two above experiments, which generate augmented images and
append them directly to the training set, we generate the augmented
images in each mini-batch and calculate the local/global regularization
terms. $n^{S}=n^{T}$ are set qual to 2, $L=15$ iterations, step
size $\eta=0.007$. The initial global and local trade-off are $3.10^{-5}$
and $50$, these parameters are is adjusted by $\frac{\mathrm{iter}}{\mathrm{\#num\_iter}}$
in $\mathrm{iter}$-th iteration. We train AlexNet for 45.000 iterations
with SGD optimizer and $10^{-3}$ learning rate.
\end{itemize}
\vspace{.5cm}
\subsubsection{Datasets and Baselines }

\begin{table}
\caption{Details on the domain generalization benchmark datasets\label{tab:dg-data}}
\centering{}\resizebox{.5\columnwidth}{!}{
\begin{tabular}{c|c|c}
\toprule
Dataset & \# classes & Shape\tabularnewline
\midrule
MNIST \citep{lecun1998gradient} & 10 & 32 \texttimes{} 32\tabularnewline
SVHN \citep{netzer2011reading} & 10 & 32 \texttimes{} 32\tabularnewline
MNIST-M \citep{ganin2015unsupervised} & 10 & 32 \texttimes{} 32\tabularnewline
SYN \citep{ganin2015unsupervised} & 10 & 32 \texttimes{} 32\tabularnewline
USPS \citep{denker1989neural} & 10 & 32 \texttimes{} 32\tabularnewline
CIFAR-10-C \citep{hendrycks2019robustness} & 15 & 3 \texttimes{} 32 \texttimes{} 32\tabularnewline
CIFAR-100-C \citep{hendrycks2019robustness} & 15 & 3 \texttimes{} 32 \texttimes{} 32\tabularnewline
PACS \citep{li2017deeper} & 7 & 3 \texttimes{} 224 \texttimes{} 224\tabularnewline
\bottomrule
\end{tabular}}
\end{table}

We present the details on each dataset used in domain generalization
experiments in Table. \ref{tab:dg-data}. Digits datasets: MNIST \citep{lecun1998gradient},
SVHN \citep{netzer2011reading}, MNIST-M \citep{ganin2015unsupervised},
SYN \citep{ganin2015unsupervised}, USPS \citep{denker1989neural}
- each contains 10 classes from $0-9$, which are resized to $32\times32$
images in our experiment. CIFAR-10-C \citep{hendrycks2019robustness},
and CIFAR-100-C \citep{hendrycks2019robustness} consist of corrupted
images from the original CIFAR \citep{krizhevsky2009learning} datasets
with 15 corruptions types applied. In terms of multi-source domain
generalization, we test our proposed model on PACS dataset \citep{li2017deeper},
which includes $3\times224\times224$ images from four different datasets,
including Photo, Art painting, Cartoon, and Sketch.

In the digits experiment, 10000 images are sellected from MNIST dataset
as the training set for the source domain and the other four data
sets as the target domains: SVHN , MNIST-M, SYN , USPS. We compare
our method with the following baselines: (i) Empirical Risk Minimization
(ERM), (ii) PAR \citep{NEURIPS2019_3eefceb8}, (iii) ADA \citep{volpi2018generalizing}
and (iv) ME-ADA \citep{zhao_maximum}. For a fair comparison, we did
not use any data augmentation in this digits experiment, all the samples
are considered as RGB images (we duplicate the channels if they are
grayscale images).

\subsubsection{Experimental Results}

\begin{table*}[!ht]
\caption{Average classification accuracy (\%) on MNIST benchmark, we first
train the LeNet5 \citep{lecun1989backpropagation} architecture on
MNIST then evaluate on SVHN, MNIST-M, SYN, USPS\label{tab:mnist-dg}.
We repeat experiment for 10 times and report the mean value and standard
deviation.}

\centering{}\resizebox{.8\columnwidth}{!}{
\begin{tabular}{c|c|c|c|c|c}
\toprule 
 & SVHN & MNIST-M & SYN & USPS & Average\tabularnewline
\midrule
Standard (ERM) & 31.95\textpm{} 1.91 & 55.96\textpm{} 1.39 & 43.85\textpm{} 1.27 & 79.92\textpm{} 0.98 & 52.92\textpm{} 0.98\tabularnewline
PAR & 36.08\textpm{} 1.27 & 61.16\textpm{} 0.21 & 45.48\textpm{} 0.35 & 79.95\textpm{} 1.18 & 55.67 \textpm{} 0.33\tabularnewline
ADA & \multicolumn{1}{c|}{35.70 \textpm{} 2.00} & 58.65\textpm{} 1.72 & 47.18\textpm{} 0.61 & 80.40\textpm{} 1.70 & 55.48\textpm{} 0.74\tabularnewline
ME-ADA & 42.00\textpm{} 1.74 & 63.98\textpm{} 1.82 & 49.80\textpm{} 1.74 & 79.10\textpm{} 1.03 & 58.72\textpm{} 1.12\tabularnewline
\midrule
GLOT-DR n=1 & 42.70 \textpm{} 1.03 & 67.72 \textpm{} 0.63 & \textbf{50.53 \textpm{} 0.88} & 82.32 \textpm{} 0.63 & 60.82 \textpm{} 0.79\tabularnewline
GLOT-DR n=2 & 42.35 \textpm{} 1.44 & 67.95 \textpm{} 0.56 & \textbf{50.53 \textpm{} 0.99} & 82.33 \textpm{} 0.61 & 60.81\textpm{} 0.90\tabularnewline
GLOT-DR n=4 & \textbf{43.10 \textpm{} 1.16} & \textbf{68.44 \textpm{} 0.46} & 50.49 \textpm{} 1.04 & \textbf{82.48 \textpm{} 0.51} & \textbf{61.13 \textpm{} 0.79}\tabularnewline
\bottomrule
\end{tabular}}
\end{table*}

Table. \ref{tab:mnist-dg} shows that our model achieves the highest
average accuracy compared to the other baselines for all values of
$n^{S}=n^{T}=n\in\left\{ 1,2,4\right\} $, with the highest overall
score when $n=4$. In particular, we observe the highest improvement
in MNIST-M target domain of $\thickapprox5\%$, and $\thickapprox2.5\%$
overall. Our GLOT-DR also exhibits more consistent with smaller variation
in terms of accuracy between runs compared to the second-best method,
$(0.79\%-1.12\%)$. 

\subsection{Experiments for DA}

\subsubsection{Network architectures and Hyperparameters}

The ResNet50 \citep{he2016deep} architecture pretrained on ImageNet,
followed by a two fully connected layers classifier. is the same as
that of the previous work. We evaluate GLOT-DR on the standard object
image classification benchmarks in domain adaptation: Office-31 and
ImageCLEF-DA. The proposed method is employed on the latent space,
trade-off parameters for global and local terms are set equal to $0.02$
and $5$ throughout all the DA experiments. We train the ResNet50
model for 20000 steps with batch size of 36, following the standard
protocols in \citep{long2017conditional}, with data augmentation techniques
like random flipping and cropping.
\subsubsection{Dataset}

The Office-31 \citep{saenko2010adapting} dataset consists of $4,110$ images, divide into $31$ classes from three domains as presented in the main paper, we conduct one more experiment on another
dataset: ImageCLEF-DA, containing 12 categories from three public
datasets: Caltech-256 (C), ImageNet ILSVRC 2012 (I) and Pascal VOC
2012 (P). Each of these domains includes 50 images per class and 600
in total, which were resized to $3\times224\times224$ in our experiment.
We evaluate all baselines in 6 adaptation scenarios as in previous studies: DAN \citep{pmlr-v37-long15},
DANN \citep{ganin2016domain}, JAN
\citep{long2017deep}, 
CDAN \citep{long2017conditional}, 
and ETD \citep{li2020enhanced}.

\subsubsection{Experimental Results}

As reported in Table. \ref{tab:clef}, the GLOT-DR approach outperforms
the comparison methods on nearly all settings, except the pairs of
I\textrightarrow P and C\textrightarrow I, where our scores are equal to ETD \citep{li2020enhanced}.
Our proposed method achieves $90.4\%$ average accuracy overall, which is the highest compared to all baselines.

\begin{table}[!ht]
\begin{centering}
\caption{Accuracy (\%) on ImageCLEF-DA of ResNet50 model \citep{he2016deep}
in unsupervised domain adaptation methods with results of related work are from original papers. \label{tab:clef}}
\par\end{centering}
\centering{}\resizebox{.6\columnwidth}{!}{
\begin{tabular}{c|cccccc|c}
\toprule
 & I\textrightarrow P & P\textrightarrow I & I\textrightarrow C & C\textrightarrow I & C\textrightarrow P & P\textrightarrow C & Avg\tabularnewline
\midrule 
ResNet & 74.8 & 83.9 & 91.5 & 78.0 & 65.5 & 91.3 & 80.7\tabularnewline
DAN & 74.8 & 83.9 & 91.5 & 78.0 & 65.5 & 91.3 & 80.7\tabularnewline
DANN & 75.0 & 86.0 & 96.2 & 87.0 & 74.3 & 91.5 & 85.0\tabularnewline
JAN & 76.8 & 88.4 & 94.8 & 89.5 & 74.2 & 91.7 & 85.8\tabularnewline
CDAN & 76.7 & 90.6 & 97.0 & 90.5 & 74.5 & 93.5 & 87.1\tabularnewline
ETD & \textbf{81.0} & 91.7 & 97.9 & \textbf{93.3} & 79.5 & 95.0 & 89.7\tabularnewline
\midrule
GLOT-DR & \textbf{81.0}  & \textbf{93.8}  & \textbf{98.0}  & \textbf{93.3 } & \textbf{79.7 } & \textbf{96.3 } & \textbf{90.4 }\tabularnewline
\bottomrule
\end{tabular}}
\end{table}

\begin{figure}[!ht]
\centering{\includegraphics[width=.8\columnwidth]{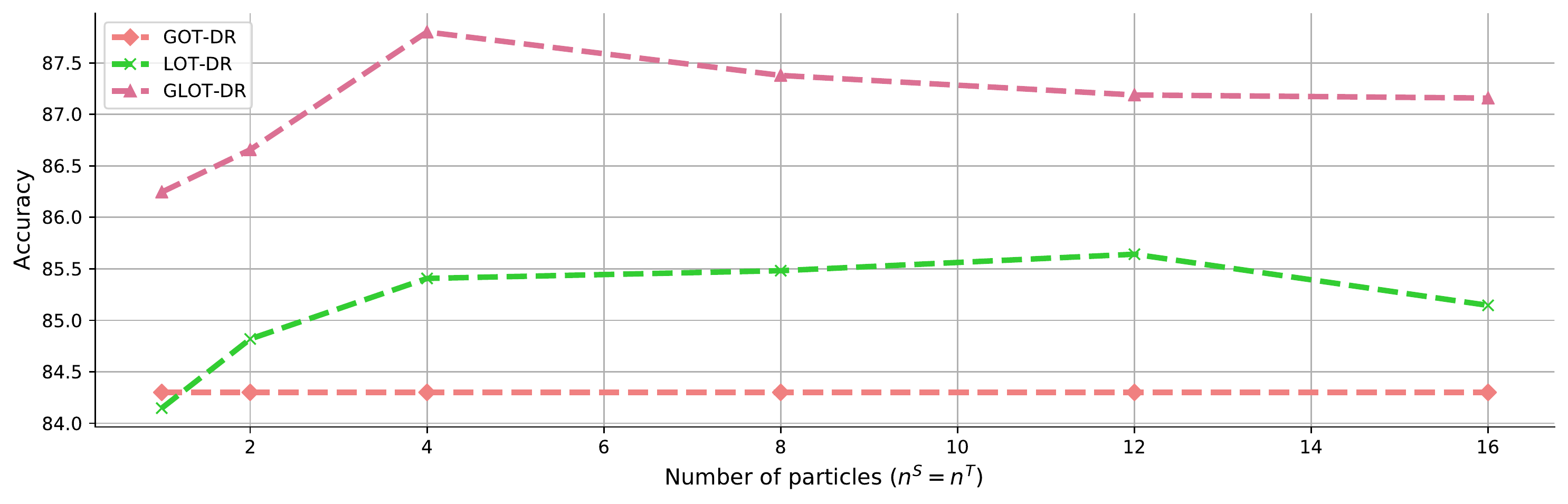}
\caption{Classification accuracy (\%) on  on Office-31 \citep{saenko2010adapting} of ResNet50 \citep{he2016deep}
model when varying the number of generated examples
sampled from Project SVGD Algorithm.\ref{alg:svgd}\label{fig:office-31-particles}. }}
\end{figure}

Up till now, we have almost finished the needed experiments to examine
the effectiveness of our method on domain adaptation. In this ultimate
experiment, we illustrate the strength of our proposed regularization
technique by varying the number of generated adversarial examples
(i.e. $n^{S}$and $n^{T}$) from 1 to 16. Results are presented in
Figure \ref{fig:office-31-particles}, where we perform extensive experiment
via comparing GLOT-DR against its variants on different values of
$n^{S},n^{T}$. It can be easily seen that, increasing the number
of generated samples can consistently improves the performance in
both LOT-DR and GLOT-DR (note that in GOT-DR there is no local regularization
term involved, thus there is no difference between different number
of samples). Setting $n^{S}=n^{T}\geq2$ helps LOT-DR surpass the
performance of GOT-DR.

\subsection{Experiments for SSL}

\subsubsection{Network architectures and Hyperparameters}

In the semi supervised learning experiment, our main competitor is
Virtual Adversarial Training (VAT) \citep{VAT}, we thus replicate
their Conv-Large\footnote{$\mathrm{LReLU}$ indicates the Leaky ReLU \citep{maas2013rectifier}
activation function with the negative slope equal to 0.1.} architecture as:
\begin{align*}
 & \mathrm{32\times32\,RGB\,image\shortrightarrow3\times3\,conv.128\,LReLU}\\
 & \mathrm{\shortrightarrow3\times3\,conv.128\,LReLU\shortrightarrow3\times3\,conv.128\,LReLU}\\
 & \mathrm{\shortrightarrow2\times2\,MaxPool,stride\,2\shortrightarrow Dropout(0.5)}\\
 & \mathrm{\shortrightarrow3\times3\,conv.256\,LReLU\shortrightarrow3\times3\,conv.256\,LReLU}\\
 & \mathrm{\shortrightarrow3\times3\,conv.256\,LReLU\shortrightarrow2\times2\,MaxPool,stride\,2}\\
 & \mathrm{\mathrm{\shortrightarrow Dropout(0.5)\shortrightarrow3\times3\;conv.512\,LReLU}}\\
 & \mathrm{\shortrightarrow1\times1\,conv.256\,LReLU\shortrightarrow1\times1\,conv.128\,LReLU}\\
 & \mathrm{\shortrightarrow Global\,Average\,Pool,6\times6\rightarrow1\times1\shortrightarrow FC_{128\times10}}
\end{align*}

We train the Conv-Large network in 600 epochs with batch size of 128
using SGD optimizer and cosine annealing learning rate scheduler \citep{loshchilov2016sgdr}.
The global and local trade-off parameters are ajusted by exponential
rampup from \citep{samuli2017temporal}:
\[
\tau=\begin{cases}
\exp^{-5(1-\mathrm{\frac{\mathrm{epoch}}{ramup\,length}})^{2}} & \mathrm{epoch<rampup\,length}\\
1 & \mathrm{otherwise}
\end{cases}
\]

with $\mathrm{rampup\,length=30}$ and initial trade-off for global
and local terms are $0.1$ and $10$, respectively.

\subsubsection{Experimental Results}

\begin{figure}[!ht]
\centering
\includegraphics[width=1\columnwidth]{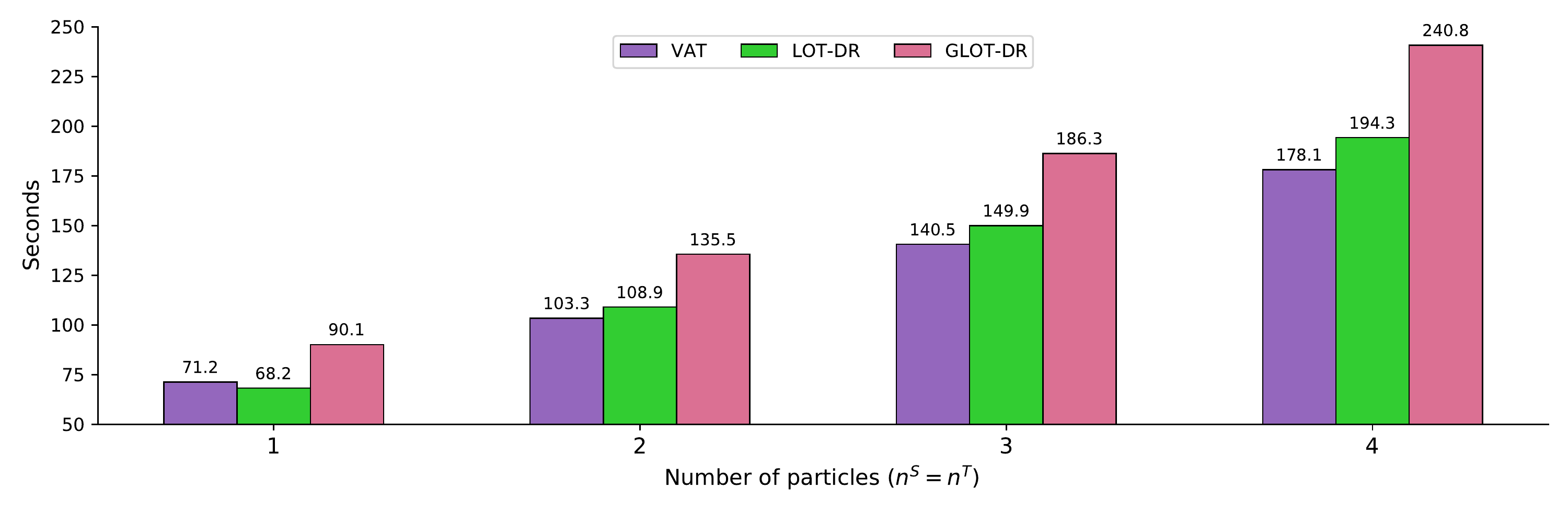}
\caption{Running time of our proposed approach on: Intel(R) Xeon(R) CPU @ 2.00GHz
CPU and Tesla P100 16GB VRAM GPU\label{fig:runtime}. Results are
averaged over 3 runs.}

\end{figure}

In this section, we compare the training time in section 4.3
of LOT-DR and GLOT-DR against VAT in a single epoch. We repeat this
process several times to get the average results, which are plotted
in Figure \ref{fig:runtime}. While VAT and LOT-DR run in almost equivalent
time for all values of generated examples, GLOT-DR requires approximately
$25\%$ extra running time. Note that this is worthy because of the
superior performance and great flexibility it brings on different
scenarios.

Furthermore, we also compare our proposed GLOT-DR with VAT \citep{VAT} and Nguyen-Duc, et al. \citep{nguyen2022particle} in the SSL scenario, utilizing the protocol from table 1 from their paper. As can be seen from Table \ref{tab:ssl}, our method is still better in all experiments ($>1\%$), especially when the number of particles $n = 8$, it outperforms all baselines by large margins.

\begin{table}[!ht]
\centering
\caption{Semi-supervised learning on Conv-Large backbone. \label{tab:ssl}}
\resizebox{.5\columnwidth}{!}{
\begin{tabular}{c|cccc}
\toprule
n particle(s) & 1 & 2 & 4 & 8 \tabularnewline
\midrule
VAT &  0.8601 &  0.8611 &  0.858 &  0.856 \tabularnewline
Nguyen-Duc, et al. & 0.867 & 0.876 & 0.883 & 0.872 \tabularnewline
\midrule 
GLOT-DR & \textbf{0.881} & \textbf{0.888} & \textbf{0.892} & \textbf{0.894} \tabularnewline
\bottomrule
\end{tabular}}
\end{table}

\subsection{Experiments for AML}

\subsubsection{General setting }

We follow the setting in \citep{pang2020bag} for the experiment on
adversarial machine learning domain. Specifically, the experiment
has been conducted on CIFAR-10 dataset with ResNet18 architecture.
All models have been trained with 110 epochs with SGD optimizer with
momentum 0.9, weight decay $5\times10^{-4}$. The initial learning
rate is 0.1 and reduce at epoch 100-th and 105-th with rate 0.1 as
mentioned in \citep{pang2020bag}. 

\subsubsection{Attack setting }

We use different SOTA attacks to evaluate the defense methods including:
(1) PGD attack \citep{madry2017towards} which is a gradient based
attack with parameter $\{k=200,\epsilon=8/255,\eta=2/255\}$ where
$k$ is the number of attack iterations, $\epsilon$ is the perturbation
boundary and $\eta$ is the step size of each iteration. (2) Auto-Attack
(AA) \citep{croce2020reliable} which is an ensemble methods of four
different attacks. We use standard version with $\epsilon=8/255$.
(3) B\&B attack \citep{brendel2019accurate} which is a decision based
attack. Following \citep{tramer2020adaptive}, we initialized with
the PGD attack with $k=20,\epsilon=8/255,\eta=2/255$ then apply B\&B
attack with 200 steps. We use $L_{\infty}$ for measuring the perturbation
size and we use the full test set of 10k samples of the CIFAR-10 dataset
in all experiments. 

\subsubsection{Baseline setting}

We compare our method with PGD-AT \citep{madry2017towards} and TRADES
\citep{zhang2019theoretically} which are two well-known defense methods
in AML. PGD-AT seeks the most violating examples that maximize the
loss w.r.t. the true hard-label $\mathcal{L}_{CE}(h_{\theta}(x_{a}),y)$
while TRADES seeks the most divergent examples by maximizing the KL-divergence
w.r.t. the current prediction (as consider as a soft-label) $\mathcal{L}_{KL}\left(h_{\theta}\left(x_{a}\right)\parallel h_{\theta}\left(x\right)\right)$.
To be fair comparison, we use the same training setting for all methods,
and succesfully reproduce performance of PGD-AT and TRADES as reported
in \citep{pang2020bag}. We also compare with adversarial distributional
training \citep{deng2020adversarial} (ADT-EXP and ADT-EXPAM) which
assume that the adversarial distribution explicitly follows normal
distribution.

\end{document}